\documentclass[journal]{IEEEtran}

\usepackage{amsmath} 
\usepackage{verbatim}
\usepackage[pdf]{pstricks}
\usepackage{tikz}
\usepackage{cleveref}
\usepackage{breqn}
\usepackage{booktabs} 
\usepackage[utf8]{inputenc}
\usepackage{algpseudocode}
\usepackage{algorithm}
\usepackage{xcolor}
\usepackage{pgfplots}
\usepackage{graphicx}
\usepackage{caption}
\usepackage{subcaption}
\usepgfplotslibrary{groupplots}
\usetikzlibrary{fit,positioning}
\usetikzlibrary{patterns}
\usetikzlibrary{bayesnet}
\usetikzlibrary{external}
\ifCLASSINFOpdf
\else
\fi
\newcommand{\vect}[1]{\boldsymbol{#1}}
\newcommand{\matr}[1]{\boldsymbol{#1}}
\newcommand{\trans}[1]{{#1}^{\top}}
\newcommand{\given}[0]{\, | \,}
\newcommand{\normal}[3]{\mathcal{N}\left({#1} \, \left| \, {#2},{#3}\right.\right)}
\newcommand{\invwish}[3]{\mathcal{W}^{-1}\left({#1} \given {#2},{#3}\right)}
\newcommand{\niw}[5]{\mathrm{NIW}\left({#1} \given {#2},{#3},{#4},{#5}\right)}
\newcommand{\Tau}{\mathrm{T}}

\newcommand{\diff}[1]{d{#1}}

\DeclareMathOperator*{\argmax}{arg\,max}

\DeclareMathOperator{\blockdiag}{blockdiag}
\DeclareMathOperator{\train}{train}
\DeclareMathOperator{\evaluate}{evaluate}

\newcommand{\wmean}[1]{\overline{\vect{w}}_{#1}}
\newcommand{\wpoint}[1]{\hat{\vect{w}}_{#1}}
\newcommand{\wcov}[1]{\matr{S}_{\omega}^{#1}}
\newcommand{\swmle}[0]{\matr{\Sigma}_\omega^{*}}
\newcommand{\muwmle}[0]{\vect{\mu}_\omega^{*}}

\newcommand{\ytdes}[0]{\vect{y}_t^*}

\newcommand{\katmomp}[0]{\textit{Offline MoMP}}
\newcommand{\seboffline}{\textit{Offline ProMP}}
\newcommand{\sebmle}{\textit{MLE ProMP}}
\newcommand{\sebmap}{\textit{MAP ProMP}}

\newlength\figureheight
\newlength\figurewidth
\newlength\axislabelsep

\newcommand{\rev}[1]{{#1}}

\newcommand{\alexp}[0]{LSM}

\begin{document}
%
\title{Adaptation and Robust Learning\\ of Probabilistic Movement Primitives}
%
%
%

\author{Sebastian~Gomez-Gonzalez, 
        Gerhard~Neumann, 
        Bernhard~Schölkopf, 
        and~Jan~Peters, 
\thanks{S. Gomez-Gonzalez, B. Schölkopf and J. Peters are with the Department
of Intelligent Systems, Max Planck Institute, Tübingen, Germany}
\thanks{J. Peters and G. Neumann are with TU-Darmstadt, Germany.}
\thanks{This work has been submitted to the IEEE for possible publication. 
Copyright may be transferred without notice.}}

%
%

\markboth{IEEE Transaction on Robotics,~Vol.~X, No.~Y, July~2017}%
{Gomez-Gonzalez \MakeLowercase{\textit{et al.}}: 
Adaptation and Robust Learning of Probabilistic Movement Primitives}
%



\maketitle

\begin{abstract}
  Probabilistic representations of movement primitives open important new possibilities
  for machine learning in robotics. These representations are able to capture the variability
  of the demonstrations from a teacher as a probability distribution over trajectories, providing a sensible
  region of exploration and the ability to adapt to changes in the robot environment.
  However, to be able to capture variability and correlations between different joints,
  a probabilistic movement primitive requires the estimation of a larger number of parameters 
  compared to their deterministic counterparts, that focus on modeling only the mean behavior.
  In this paper, we make use of prior distributions over the parameters of a probabilistic
  movement primitive to make robust estimates of the parameters with few training instances. 
  In addition, we introduce general purpose operators to adapt movement primitives in 
  joint and task space.
  The proposed training method and adaptation operators are tested in a coffee preparation
  and in robot table tennis task. In the coffee preparation task we evaluate the generalization
  performance to changes in the location of the coffee grinder and brewing chamber in a target
  area, achieving the desired behavior after only two demonstrations.
  In the table tennis task we evaluate the hit and return rates, outperforming previous
  approaches while using fewer task specific heuristics.
\end{abstract}

\begin{IEEEkeywords}
Robot Learning, Robot Motion
\end{IEEEkeywords}

%
\IEEEpeerreviewmaketitle

\section{Introduction}
%
%
%
%

\IEEEPARstart{T}{echniques} that can learn motor behavior from human demonstrations and 
reproduce the learned behavior in a robotic system have the potential to generalize better 
to different tasks. Multiple models have been proposed to represent complex
behavior as a sequence of simpler movements typically known as movement primitives.
A movement primitive framework should provide operators to learn primitives from demonstrations,
adapt them to achieve different goals and execute them in a sequence on a robotic system. 

Deterministic Movement Primitive frameworks have been used successfully for a variety of robotic 
tasks including locomotion~\cite{dmp_locomotion}, grasping~\cite{ude2010_dmp_gen}, ball in a 
cup~\cite{kober2014learning} and pancake flipping~\cite{dmp_pancake}. 
However, deterministic representations capture only the mean
behavior of the demonstrations of the teacher. The variability in the demonstrations 
is not captured nor used. 

In biological systems, variability seems to be characteristic of all behavior, even in 
the most skilled and seemingly automated performance~\cite{muller2009motor}. Thus, a 
movement primitive representation that captures variance in the demonstrated behavior 
has the potential to model the human teacher better. 
For a task like table tennis, the variability of the teacher is partially a response 
to the changes in ball trajectory. Therefore, approaches that capture it have the potential 
to adapt better to diverse ball trajectories. At the same time, the variability of the 
teacher can be used to define a region of sensible exploration for a robotic system.

Probabilistic approaches can naturally capture variability using a probability
distribution. Some probabilistic representations of movement primitives focus
on learning a distribution over demonstrated states using Gaussian Mixture models
or Hidden Markov models~\cite{calinon2007learning}\cite{billard2008robot}. Subsequently 
using the log-likelihood as cost function to reproduce the learned movement using an 
optimal control method~\cite{calinon2014task}\cite{medina2012risk}. 

Other probabilistic representations focus on learning a distribution over robot 
trajectories directly. Some approaches represent trajectories as functions of time
and the distribution over these trajectories using parametric~\cite{proMP} or
non-parametric~\cite{alvarez2009latent} approaches. The trajectories
can also be represented with recursive probability distributions, using latent state 
space models~\cite{chiappa2009using}.

In this paper, we build on top of a probabilistic representation introduced in~\cite{proMP}
called Probabilistic Movement Primitives (ProMPs). In this probabilistic formulation of 
movement primitives, a movement primitive is represented
as a probability distribution over robot trajectories. Different realizations of the same 
movement primitive are assumed to be independent samples from the distribution over
trajectories.

\begin{figure}[b]
  \centering
  \includegraphics[width=7cm]{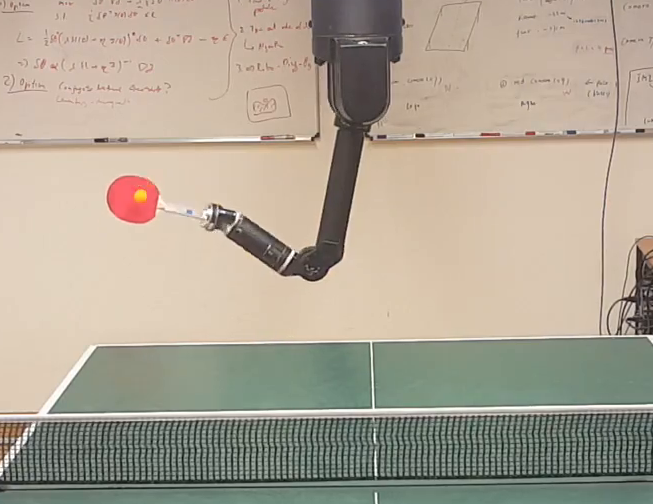}
  \caption{
    Robot table tennis setup used to evaluate the proposed methods. The ball
    is tracked using four cameras attached to the ceiling. The robot arm is
    a Barrett WAM capable of high speed motion, with seven degrees of freedom
    like a human arm.
  }
  \label{fig:robot}
\end{figure}

ProMPs have typically more parameters than non-probabilistic representations. These extra 
parameters are used to capture the variability of the movements executed by the teacher 
and the correlations between different degrees of freedom of the robot. 
In this paper, we use prior distributions over the ProMP parameters to make robust 
estimates with few demonstrations. The influence of the prior distribution decreases
as more training data becomes available, converging to the maximum likelihood estimates.

This paper also presents general purpose operators to adapt a ProMP to have a desired 
joint or task space configuration at a certain time. By joint space we refer to the joint angles 
and velocities of the robot, and by task space we refer to the world coordinate position
and velocity of the end effector of the robot. 


The proposed method to learn the movement primitive and the operators to adapt the
movement primitives in task and joint space are evaluated with synthetic data, in a 
robot table tennis task and a robot assisted coffee preparation task. Figure~\ref{fig:robot} 
shows the robot table tennis setup used in the experiments. The results obtained with
the presented method are compared with previous work on robot table tennis. 
The proposed approach outperforms previous robot table tennis approaches using 
less task specific heuristics. Examples of task specific heuristics used for robot 
table tennis in previous approaches include using a Virtual Hitting 
Planes~\cite{mulling2011biomimetic} and computing optimal racket velocity and
orientation at hitting time to send balls to the opponent side of the 
table~\cite{mulling2013learning}.
The presented approach does not compute racket orientations or velocities to return
balls to the opponent's court. The training data used to learn the movement primitives
was built using only successful human demonstrations. The robot was able to learn
the behavior required to successfully return balls to the opponent side of the table
from the human demonstrations.

We use the pouring coffee task to evaluate the generalization performance 
of the presented method as a function of the number
of training instances by changing the position of the coffee grinder and the brewing chamber.
The robot manages to pour successfully on the selected testing area after two training
demonstrations, suggesting that the presented prior is a sensible choice for this task.
Finally, the fact that the presented approach can be used for two robot tasks as different
as table tennis and coffee pouring without any changes suggests it has the potential to 
perform well in several other robot applications.

\section{Robust Learning of Probabilistic Movement Primitives}
\label{sec:methods}

\begin{figure}
  \setlength\figurewidth{7cm}
  \setlength\figureheight{4cm}
  \centering
  \tikzsetnextfilename{promp_graphmod}
  \resizebox{\figurewidth}{\figureheight} {
\begin{tikzpicture}
\tikzstyle{main}=[circle, minimum size = 10mm, thick, draw =black!80, node distance = 16mm]
\tikzstyle{connect}=[-latex, thick]
\tikzstyle{box}=[rectangle, draw=black!100]
  \node[main, fill = white!100] (mu_w) [label=below:$\vect{\mu}_\omega$]{ };
  \node[main, fill = white!100] (Sigma_w) [below=of mu_w,label=below:$\matr{\Sigma}_\omega$] { };
  \node[main] (omega) [right=of {$(mu_w)!0.5!(Sigma_w)$},label=below:$\vect{\omega}_{n}$] { };
  \node[main, fill = black!25] (y) [right=of omega,label=below:$\vect{y}_{nt}$] {};
  \path (mu_w) edge [connect] (omega)
        (omega) edge [connect] (y);
  \path (Sigma_w) edge [connect] (omega);
  \node[rectangle, inner sep=0.5mm, fit= (y),label=below right:$\Tau_{n}$, xshift=-1.0mm] {}; 
  \node[rectangle, inner sep=5.5mm,draw=black!100, fit= (y)] (rep_t) {};
  \node[rectangle, fit= (omega) (rep_t),label=below right:$N$, xshift=8.5mm, yshift=1mm] {}; 
  \node[rectangle, inner sep=5mm, draw=black!100, fit = (rep_t) (omega)] {};
\end{tikzpicture}
}
  \caption {
    Probabilistic Movement Primitive graphical model. 
    The joint state~$\vect{y}_{nt}$ is generated from the compact representation of a 
    trajectory~$\vect{\omega}_n$ using~\eqref{eq:promp:proMP}.
    The mean behavior of the different trajectories is represented by the 
    variable~$\vect{\mu}_\omega$. The variability of the teacher and the correlations between
    different joints are represented by~$\matr{\Sigma}_\omega$. Each 
    trajectory~$\vect{\omega}_n$ is generated using~\eqref{eq:promp:distw}. The number of 
    trials is denoted by~$N$, and the number of time steps of the trial $n$ is 
    denoted by~$\Tau_{n}$.
  }
  \label{fig:promp:model}
\end{figure}
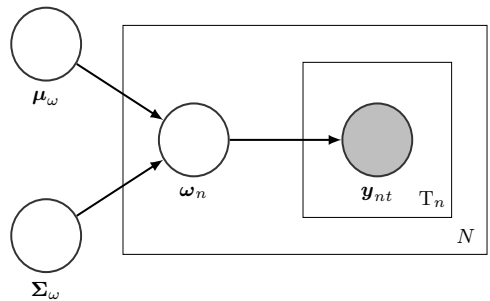

Probabilistic movement primitives (ProMPs) are probability distributions used
to represent motion trajectories~\cite{proMP}. A trajectory~$\tau = \{\vect{y}_t\}_{t=1}^T$, 
can be represented as positions or joint angles at different moments in time.
In this paper, we assume that~$\vect{y}_t$ is a $D$ dimensional vector that
represents the joint measurement at time $t$ of a robotic system with $D$
degrees of freedom.

First, let us introduce a
variable~$\vect{\omega}=\trans{[\trans{\vect{\omega}_1},\dots,\trans{\vect{\omega}_D}]}$ 
that encodes compactly
a single robot trajectory, and consists of the concatenation of~$D$ 
weight vectors~$\vect{\omega}_d$ that represent the trajectory of each of the 
degrees of freedom of the robot, indexed by~$d$. 

Given a trajectory realization represented by~$\vect{\omega}$, the joint state at 
time $t$ is computed as
\begin{equation*}
  \vect{y}_t = \trans{[
    \trans{\vect{\phi}_1(t)}\vect{\omega}_1, 
    \cdots, 
    \trans{\vect{\phi}_D(t)}\vect{\omega}_D
  ]} + \vect{\epsilon}_y,
\end{equation*}
where the vector~$\vect{\phi}_d(t)$ is a computed from a set of time dependent basis functions,
and~$\vect{\epsilon}_y$ is Gaussian white noise. To obtain smooth trajectories, the basis
functions need to be smooth. In this paper we use radial basis functions (RBF), 
polynomial basis functions and a combination of both. 
The number and type of basis functions to use is a design choice. 
Each degree of freedom could have a different
number of basis functions, but for simplicity we assume every degree of freedom 
uses~$K$ basis functions. 

The distribution over the values of the joint state at time~$t$, can be written as
\begin{equation}
  p(\vect{y}_{t}|\vect{\omega}) = \normal{\vect{y}_{t}}{\matr{\Phi}_{t}\vect{\omega}}{\matr{\Sigma}_y},
  \label{eq:promp:proMP}
\end{equation}
where~$\matr{\Phi}_t$ is a~$D \times KD$ matrix used to write the distribution over~$\vect{y}_t$ in
vectorized form, and is defined as
\[
  \matr{\Phi}_t = 
  \begin{pmatrix}
    \vect{\phi}_1(t) & \cdots & \vect{0} \\
    \vdots & \ddots & \vdots \\
    \vect{0} & \cdots & \vect{\phi}_D(t)
  \end{pmatrix}.
\]
Different realizations of a movement primitive are assumed to have different values 
for~$\vect{\omega}$. In this model, a particular realization~$n$ represented by~$\vect{\omega}_n$ 
is assumed to be sampled from
\begin{equation}
  p(\vect{\omega}_n|\vect{\theta}_\omega) = \normal{
    \vect{\omega}_n
  }{
    \vect{\mu}_{\omega}
  }{
    \matr{\Sigma}_{\omega}
  },
  \label{eq:promp:distw}
\end{equation}
where~$\vect{\theta}_\omega = \{\vect{\mu}_\omega, \matr{\Sigma}_\omega\}$ is a set of
parameters that capture the similarities and differences of different realizations
of the movement primitive. In the rest of this section, we drop the index~$n$ 
from~$\vect{\omega}_n$ for notational simplicity. 

\begin{figure}
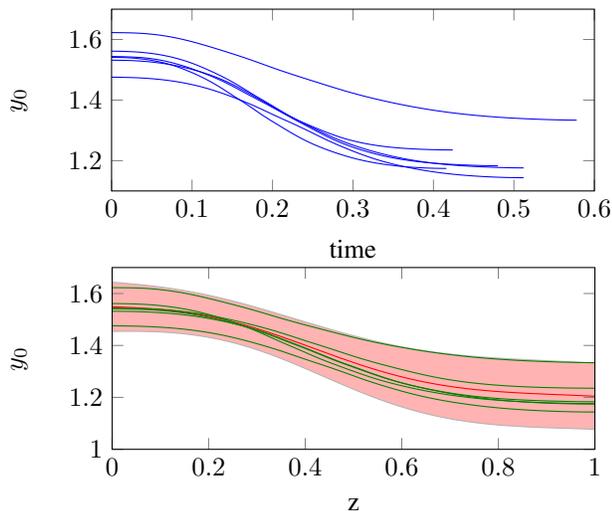

  \setlength\figurewidth{8cm}
  \setlength\figureheight{4cm}
  \setlength\axislabelsep{0mm}
  \centering
  \tikzsetnextfilename{q0samples}
  \input{fig/q0_samples.tex}
  \tikzsetnextfilename{q0}
  \input{fig/q0.tex}
  \caption {
    Demonstrated trajectories and learned distribution for the first 
    degree of freedom of Barrett WAM robot arm. The joint value $y_0$ 
    corresponds to the shoulder yaw recorded in radians as a function of time. 
    Different trajectories of a table tennis forehand motion are demonstrated by the 
    human teacher. These trajectories are depicted in blue and have different durations.
    The duration of each trajectory is normalized to one to achieve duration invariance,
    the time invariant trajectories are depicted in green.
    The learned distribution 
    is depicted in red. The shaded area corresponds to two standard deviations. Note
    that the model captures the mean behavior and the variability of the teacher
    at different points in time.
  }
  \label{fig:tt:joint_dist}
\end{figure}

Let us write the distribution~$p(\vect{\omega}|\vect{\theta}_\omega)$ 
decomposing the~$KD \times 1$ vector~$\vect{\mu}_\omega$ and the~$KD \times KD$ 
matrix~$\matr{\Sigma}_\omega$ in the components corresponding to each degree of freedom,
\begin{equation*}
  \normal{
    \begin{pmatrix}
      \vect{\omega}_1 \\
      \vdots \\
      \vect{\omega}_D \\
    \end{pmatrix}
  }{
    \begin{pmatrix}
      \vect{\mu}_{\omega}^1 \\
      \vdots \\
      \vect{\mu}_{\omega}^D \\
    \end{pmatrix}
  }{
    \begin{pmatrix}
      \matr{\Sigma}_{\omega}^{(1,1)} & \cdots & \matr{\Sigma}_{\omega}^{(1,D)} \\
      \vdots & \ddots & \vdots \\
      \matr{\Sigma}_{\omega}^{(D,1)} & \cdots & \matr{\Sigma}_{\omega}^{(D,D)} \\
    \end{pmatrix}
  }.
  \label{eq:promp:distw_dof}
\end{equation*}
Note that the mean behavior for the degree of freedom~$d$ is captured by the~$K \times 1$ 
vector~$\vect{\mu}_{\omega}^d$ and the variability by the~$K \times K$ 
matrix~$\matr{\Sigma}_{\omega}^{(d,d)}$.
The correlation between two different joints~$d_1$ and~$d_2$ is captured 
by~$\matr{\Sigma}_{\omega}^{(d_1,d_2)}$. The model can be forced to consider all the joints 
independently by forcing the matrix~$\matr{\Sigma}_\omega$ to be block diagonal.

A probabilistic graphical model is a probabilistic model for which a graph expresses
conditional independence assumptions between random variables~\cite{murphy2012machine}.
Figure~\ref{fig:promp:model} shows the graphical representation of the probabilistic 
model used to represent movement primitives. To sample a robot trajectory given
the ProMP parameters~$\vect{\mu}_\omega$ and~$\matr{\Sigma}_\omega$, a 
vector~$\vect{\omega}_n$ is sampled using~\eqref{eq:promp:distw}. Subsequently, the
new trajectory of length~$\Tau_n$ can be sampled using~\eqref{eq:promp:proMP}.
If the used basis functions are smooth, the sampled trajectories will also be smooth.

Figure~\ref{fig:tt:joint_dist} show six human demonstrations of a forehand table tennis 
striking movement and the learned probability distribution. The figure shows the value in 
radians of the shoulder yaw~$y_0$ with respect to time. The original demonstrations given
by the human teacher, depicted in blue, have different durations varying between 0.4 and 
0.6 seconds. The time of every demonstration is normalized to be between zero and one to achieve 
duration invariance using a new variable~$z=\frac{t-t0}{T}$, known as the phase 
variable~\cite{proMP}. 
The same demonstrations with respect to the phase variable are depicted in green, and the learned
distribution is depicted in red. The shaded area corresponds to two standard deviations.
The ProMP learned from the given demonstrations capture the mean behavior and the
teacher variability in different points of time.

%

\subsection{Learning from Demonstrations}

The parameters~$\vect{\theta}_\omega$ can be learned from human demonstrations.
Let us assume we have~$N$ recorded human demonstrations, and extend the notation
with an extra sub-index~$n \in \{1..N\}$ identifying each demonstration.
Thus, the variables~$\vect{y}_{nt}$ and~$\vect{\omega}_n$ represent the joint 
state of the~$n$th trial at time $t$ and the compact representation of the~$n$th 
trial respectively. The likelihood of the recorded data is given by
\begin{equation*}
  p(\matr{Y}|\vect{\theta}_\omega) = \prod_{n=1}^{N}{\int{p(\vect{\omega}_n \given \vect{\theta}_\omega)
  \prod_{t=1}^{\Tau_n}{p(\vect{y}_{nt} \given \vect{\omega}_n)} \diff{\vect{\omega}_n}} },
  \label{eq:promp:marginalLH}
\end{equation*}
where~$\matr{Y}$ is the set of values~$\vect{y}_{nt}$ for all the training instances.
Note that evaluating the likelihood requires the computation of an integral over the
hidden variables~$\vect{\omega}_n$. Although the integral in this case can be computed
in closed form, evaluating the resulting expression would cost cubic time over the
trajectory lengths~$\Tau_n$. Instead, we propose to use the expectation maximization algorithm
to optimize the likelihood or posterior distribution with linear time costs over the 
trajectory lengths~$\Tau_n$.

In~\cite{paraschos2015model}, the ProMP parameters are estimated by first making a
point estimate of the hidden variables with least squares, and subsequently finding
their empirical mean and covariance matrix as the ProMP parameters. This estimation 
procedure makes intuitive sense and avoids computing integrals. However, the authors 
did not provide a mathematical intuition of how their estimation procedure relates
to maximizing the marginal likelihood. In the Appendix~\ref{sec:rel_alex_method},
we explain in detail the estimation method introduced in~\cite{paraschos2015model},
and show that it is an special case of an approximation of the proposed EM algorithm 
to maximize the likelihood. The approximation consists of performing a single EM 
iteration and approximating the Gaussian distribution computed in the E-step with a 
Dirac delta distribution, ignoring the uncertainty over the estimates of the
hidden variables.

In previous work~\cite{gomez2016using}, the parameters were learned maximizing the
likelihood. However, maximizing the likelihood results in numerically unstable estimates
for the parameters of the ProMP unless a very large number of demonstrations is available.
In~\cite{gomez2016using}, the matrix~$\matr{\Sigma}_\omega$ is forced to be block 
diagonal to deal with the numerical problems. As a result, the ProMP parameters
could be robustly estimated, but the model becomes incapable of learning the correlation 
between the different joints of the robot arm.

In this paper, we use regularization to estimate the ProMP parameters in the form 
of a prior probability distribution~$p(\vect{\theta}_\omega)$. The posterior
distribution over the ProMP parameters is given by
\begin{equation}
  p(\vect{\theta}_\omega | \matr{Y}) \propto p(\vect{\theta}_\omega)p(\matr{Y}|\vect{\theta}_\omega).
  \label{eq:promp:train_posterior}
\end{equation}
We estimate the parameters~$\vect{\theta}_\omega$ by maximizing the posterior distribution 
of~\eqref{eq:promp:train_posterior} using the expectation maximization
algorithm. This estimator is commonly known as Maximum A Posteriori (MAP) estimate.

The pseudo-code summarizing the training procedure is presented
in Algorithm~\ref{alg:em_promp}. Lines~\ref{alg:em_promp:estep:cov} 
and~\ref{alg:em_promp:estep:mean} correspond to the E-step and 
lines~\ref{alg:em_promp:mstep:mean_mle} to~\ref{alg:em_promp:mstep:noise}
correspond to the M-step. The values~$\vect{\epsilon}_{nt} = \vect{y}_{nt} - \matr{\phi}_{nt}\wmean{n}$ 
are the residuals used to estimate the sensor noise.

\begin{algorithm}[t]
  \begin{algorithmic}[1]
    \Require Demonstration dataset containing the joint states and corresponding normalized time 
      stamps~$\matr{Y}=\{\vect{y}_{nt}, z_{nt}\}$ and the prior parameters $k_0, \vect{m}_0, v_0, \matr{S}_0$
    \Ensure The ProMP parameters~$\vect{\mu}_\omega$, $\matr{\Sigma}_\omega$, $\matr{\Sigma}_y$
    \State Compute matrices~$\matr{\phi}_{nt}=\matr{\phi}(z_{nt})$ with the basis functions~$\matr{\phi}$
    \State Compute $L = \sum_{n=1}^N{\tau_n}$
    \State Set some initial values for~$\vect{\mu}_\omega$, $\matr{\Sigma}_\omega$, $\matr{\Sigma}_y$. We
      use~$\vect{\mu}_\omega=\vect{0}$, $\matr{\Sigma}_\omega=\matr{I}$ and~$\matr{\Sigma}_y=\matr{I}$.
    \While{Not converged}
      \For{ $n \in \{1,\dots,N\}$ }
        \State $\wcov{n} \gets \left(\Sigma_\omega^{-1} + 
          \sum_{t=1}^{\tau_n}{\trans{\matr{\phi}_{nt}}\matr{\Sigma}_y^{-1}\matr{\phi}_{nt}}\right)^{-1}$ 
          \label{alg:em_promp:estep:cov}
        \State $\wmean{n} \gets \wcov{n}\left(\matr{\Sigma}_\omega^{-1}\vect{\mu}_\omega +
          \sum_{t=1}^{\tau_n}{\trans{\matr{\phi}_{nt}}\matr{\Sigma}_y^{-1}\vect{y}_{nt}}\right)$ 
          \label{alg:em_promp:estep:mean}
      \EndFor
      \State $\muwmle \gets \frac{1}{N}\left(\sum_{n=1}^N {\wmean{n}}\right)$
        \label{alg:em_promp:mstep:mean_mle}
      \State $\vect{\mu}_\omega \gets \frac{1}{N + k_0}\left( k_0 \vect{m}_0 + N \muwmle \right)$
        \label{alg:em_promp:mstep:mean}
      \State $\swmle \gets \frac{1}{N}{\sum_{n=1}^N{\left(\wcov{n} + (\wmean{n} - \vect{\mu}_\omega)\trans{(\wmean{n} - \vect{\mu}_\omega)}\right)}}$
        \label{alg:em_promp:mstep:cov_mle}
      \State $\matr{\Sigma}_\omega \gets \frac{1}{N + v_0 + KD + 1}\left[\matr{S}_0 + N\swmle\right]$
        \label{alg:em_promp:mstep:cov}
      \State $\matr{\Sigma}_y \gets \frac{1}{L}\sum_{n=1}^N{
        \sum_{t=1}^{\tau_n}{\left[
          \vect{\epsilon}_{nt}\trans{\vect{\epsilon}_{nt}} + 
        \matr{\phi}_{nt}\wcov{n}\trans{\matr{\phi}_{nt}}\right]}
      }$ \label{alg:em_promp:mstep:noise}
    \EndWhile
    \State \Return~$\vect{\mu}_\omega$, $\matr{\Sigma}_\omega$ and~$\matr{\Sigma}_y$.
  \end{algorithmic}
  \caption{Expectation Maximization algorithm to train a ProMP from demonstrations}
  \label{alg:em_promp}
\end{algorithm}

%

\subsection{Prior Distribution}

We use a Normal-Inverse-Wishart as a prior distribution over the ProMP 
parameters~$\vect{\mu}_\omega$ and~$\matr{\Sigma}_\omega$, given by
\begin{align*}
  p(\vect{\mu}_\omega,\matr{\Sigma}_\omega) &= \niw{\vect{\mu}_\omega,\matr{\Sigma}_\omega}{k_0}{\vect{m}_0}{v_0}{\matr{S}_0} \\
  &= \normal{\vect{\mu}_\omega}{\vect{m}_0}{\frac{1}{k_0}\matr{\Sigma}_\omega} \invwish{\matr{\Sigma}_\omega}{v_0}{\matr{S}_0},
\end{align*}
where~$\invwish{\matr{\Sigma}_\omega}{v_0}{\matr{S}_0}$ is an inverse Wishart distribution,
used frequently as a prior for covariance matrices. The main reason why we decided to use a
Normal-Inverse-Wishart prior for the ProMP model is because it is a conjugate prior,
resulting in closed form updates for the parameters in the EM algorithm and simplifying the
inference process. Furthermore, the parameters of this prior distribution have a simple 
interpretation. Lines~\ref{alg:em_promp:mstep:mean_mle}
and~\ref{alg:em_promp:mstep:cov_mle} compute the Maximum Likelihood estimates (MLE)~$\muwmle$ 
and~$\swmle$. Lines~\ref{alg:em_promp:mstep:mean} and~\ref{alg:em_promp:mstep:cov} compute the 
MAP estimates~$\vect{\mu}_\omega$ and~$\matr{\Sigma}_\omega$. Note that the MAP estimates
are a weighted average of the MLE estimates~$\muwmle$ and~$\swmle$ and the assumed prior
parameters for the mean~$\vect{m}_0$ and covariance~$\matr{S}_0$ respectively. In the limit of
infinite data, the MAP estimates converge to the MLE estimates.

We use a non informative prior for~$\vect{\mu}_\omega$ in our experiments
by setting~$k_0 = 0$. Note that by setting~$k_0=0$, the MAP estimate~$\vect{\mu}_\omega$
becomes the MLE estimate~$\muwmle$. If a large number of basis functions is used, a sensible 
choice for the prior parameters is to use~$\vect{m}_0=\vect{0}$ and $k_0>0$. Such a prior will
prevent large values on the estimated vector~$\vect{\mu}_\omega$, similar to the regularization
used in Ridge Regression.

\begin{figure}
  \centering
  \setlength\figurewidth{8cm}
  \setlength\figureheight{5cm}
  \setlength\axislabelsep{0mm}
  \tikzsetnextfilename{logcondsw2}
\begin{tikzpicture}

\definecolor{color1}{rgb}{1,0.498039215686275,0.0549019607843137}
\definecolor{color0}{rgb}{0.12156862745098,0.466666666666667,0.705882352941177}
\definecolor{color3}{rgb}{0.83921568627451,0.152941176470588,0.156862745098039}
\definecolor{color2}{rgb}{0.172549019607843,0.627450980392157,0.172549019607843}
\definecolor{color4}{rgb}{0.580392156862745,0.403921568627451,0.741176470588235}

\begin{axis}[
title={Log Condition Number of $\Sigma_w$},
xlabel={Instances},
ylabel={$\log{\kappa(\Sigma_w)}$},
xmin=1.0, xmax=104.95,
ymin=5.19323420630898, ymax=30.7970055973459,
width=\figurewidth,
height=\figureheight,
tick align=outside,
tick pos=left,
x grid style={lightgray!92.02614379084967!black},
y grid style={lightgray!92.02614379084967!black},
legend style={draw=white!80.0!black},
legend entries={{MAP},{MLE},{\alexp{} $\lambda=0.0$},{\alexp{} $\lambda=0.1$},{\alexp{} $\lambda=1.0$}},
legend cell align={left}
]
\addlegendimage{no markers, color0}
\addlegendimage{no markers, color1}
\addlegendimage{no markers, color2}
\addlegendimage{no markers, color3}
\addlegendimage{no markers, color4}
\addplot [semithick, color0]
table {%
1 9.13207540164585
2 8.30439101991233
3 7.75394478912242
4 7.33451313866483
5 7.10199361304911
6 6.83754405723635
7 6.80993980449257
8 6.85879748391743
9 6.79259968512425
10 6.90671365872074
11 6.50266383807775
12 6.95619519894821
13 6.74500184439421
14 6.6327999199575
15 6.85502003784865
16 6.63701329081249
17 6.58430825025209
18 6.90661057952875
19 6.5977238387892
20 6.35704199681066
21 6.6716376658144
22 6.50696494949895
23 6.7254125248832
24 6.65063391697787
25 6.84263936246003
26 6.49560066399165
27 6.62024975008634
28 6.61626234371101
29 6.65414349652979
30 6.59583032159432
31 6.61813238993394
32 6.56933820660289
33 6.4053630317408
34 6.73669790594705
35 6.71420812047454
36 6.86201856895587
37 6.82671651991612
38 6.51862419340056
39 6.74757698494401
40 6.57311665078533
41 6.75902674712151
42 6.43141894277008
43 6.46732969733098
44 6.66026916304138
45 6.77698309180274
46 6.49461400078821
47 6.7191026316082
48 6.68475676811049
49 6.6785929882278
50 6.76181318106123
51 6.70214163347979
52 6.63338954789053
53 6.71789581880134
54 6.69230174872596
55 6.67332490312661
56 6.72406900121395
57 6.86543525332209
58 6.52263725352639
59 6.63848892276133
60 6.56198604913815
61 6.62348610693718
62 6.61233720122443
63 6.75907403470285
64 6.74971308209363
65 6.87370443042132
66 6.71175357783751
67 6.67569273061272
68 6.66752140171763
69 6.64333429570902
70 6.76808325790202
71 6.64215090190416
72 6.87758055354054
73 6.65709271490995
74 6.67693725818221
75 6.61485149094542
76 6.82414258574359
77 6.87979911525448
78 6.80082840723281
79 6.81975595434711
80 6.86960200511922
81 6.73454176182252
82 6.92680058408337
83 6.7730751484202
84 6.77486539982096
85 6.76091458750647
86 6.7602874623611
87 6.71977027461335
88 6.72353075639897
89 6.67660810200778
90 6.75630317534818
91 6.81630868782453
92 6.62387738766595
93 6.73970788468543
94 6.74757136604215
95 6.70875618063958
96 6.79347564333255
97 6.79846468338832
98 6.8395805558012
99 6.86473165778155
100 6.82851485726053
};
\addplot [semithick, color1]
table {%
2 29.6331978068443
3 29.0776143206821
4 28.8662274301515
5 28.4162790400574
6 28.1268053934089
7 28.5367866374433
8 27.933936249128
9 27.7139708167204
10 27.7543805865991
11 27.3624412466025
12 27.1607283119111
13 27.0124747666363
14 27.2405152656819
15 26.8348755433784
16 26.6466734915961
17 26.2010741000501
18 25.8983032446342
19 25.9905337608245
20 26.004849437245
21 25.4250785882187
22 25.670321049466
23 25.3288948725374
24 25.2197508755816
25 24.9808448828166
26 25.1625152381117
27 24.6146152368876
28 24.3643683939588
29 24.0893113221794
30 24.0311800920377
31 23.650393352234
32 23.690389896094
33 22.9185270917414
34 23.2130973544648
35 22.1125916900283
36 13.3710037180071
37 12.6219915097433
38 11.6441934462344
39 11.1191310194151
40 10.7652862746258
41 10.3260217853605
42 10.2269769924934
43 10.0499895145309
44 9.8917665183345
45 9.53030239320366
46 9.33769491682253
47 9.25058712925422
48 9.44861369979061
49 9.00031267142751
50 9.06564931403333
51 9.0554005218485
52 9.00473840682494
53 9.11908866925677
54 8.76713395795047
55 8.80626632557608
56 8.89615010728745
57 8.67801608141308
58 8.31724789984045
59 8.54923336637813
60 8.51368701548305
61 8.50355407509855
62 8.38657880283481
63 8.39406440645791
64 8.07147946783019
65 8.48189323794304
66 8.28057631757703
67 8.19109741707229
68 8.19205939620159
69 8.1830578556096
70 8.28882233044539
71 8.26636253178598
72 8.2622742818019
73 8.12085014287586
74 7.98718882572127
75 8.21319627380038
76 8.11340486358158
77 8.0957087980735
78 8.14368967295215
79 7.93580269201964
80 8.07222407059122
81 7.99157050885219
82 7.97426272880001
83 8.08321126903707
84 8.01299956667085
85 7.87600330519484
86 7.95810083478377
87 7.88967242621222
88 7.92647122607964
89 7.79333052206889
90 7.93021428829581
91 7.97606181130243
92 7.84189962966065
93 7.98509067596321
94 7.93317808221981
95 7.88309803732631
96 7.85683300731277
97 7.9773619177785
98 7.9577017648024
99 7.73652244967335
100 7.77922164882036
};
\addplot [semithick, color2]
table {%
36 12.1621028127299
37 12.5886933937691
38 11.1486701174831
39 10.2973806727618
40 10.1014173954216
41 10.153909396001
42 9.43713626006696
43 8.94454344952783
44 9.65367016111021
45 9.55344133404598
46 8.86307863636378
47 8.53953120713199
48 8.83181526861854
49 8.78379737305108
50 8.46720042998167
51 8.48746196252309
52 8.53969352762745
53 8.4885847203917
54 8.16804172835172
55 8.02824499446815
56 8.22385980082984
57 8.26883083978679
58 8.25290390072991
59 8.02468175875981
60 8.03869915876402
61 8.26512489016039
62 8.0032007152847
63 8.11417536575652
64 7.91899319301244
65 8.06028714332175
66 7.92411494089778
67 7.9483598858816
68 7.78797317982312
69 7.55031677858995
70 7.70576593191976
71 7.7675216838788
72 7.77425823747768
73 7.74466126248238
74 7.90151001925714
75 7.67349281429359
76 7.64639798674255
77 7.75578776648974
78 7.66048334650905
79 7.67668411739842
80 7.65216302710563
81 7.42565152147946
82 7.57315490735237
83 7.62596830615569
84 7.53033375858814
85 7.72259314444171
86 7.64015661841189
87 7.48210703528814
88 7.49418578221711
89 7.40417092906166
90 7.53170273362962
91 7.45758059385966
92 7.70663297715171
93 7.57504673636169
94 7.28424176322669
95 7.62475721374771
96 7.41328829055853
97 7.63741448719314
98 7.37871872713979
99 7.47417126594599
100 7.44662208022186
};
\addplot [semithick, color3]
table {%
36 14.0783547469409
37 11.692743383208
38 11.3146235868948
39 11.3099501190096
40 10.1198127457424
41 9.82605662572069
42 9.83378962166377
43 9.38933609454273
44 9.32790268321806
45 9.34381207262802
46 9.26884520800021
47 9.18353500134981
48 8.72340583451426
49 8.4821653909308
50 8.90154468583293
51 8.53024099103227
52 8.42190215189964
53 8.73389581252304
54 8.51455378625223
55 8.38090178661986
56 8.22731658398515
57 8.10961650006604
58 8.52099468659466
59 8.1249031195506
60 8.1799704392243
61 8.08281131369801
62 7.88715380757845
63 8.08729084773265
64 7.96655087109621
65 7.89691234501728
66 7.7658172671186
67 7.98781695150678
68 7.83107869517724
69 7.89700336821032
70 7.94478715094256
71 7.81780880217057
72 7.61348642480505
73 7.72278542384649
74 7.66648953808208
75 7.68997632066044
76 7.80928477632297
77 7.84134038526869
78 7.53029841251747
79 7.61361950034993
80 7.49803186825886
81 7.45553836044605
82 7.47841845646703
83 7.52305977619612
84 7.56748349074492
85 7.62747724553857
86 7.53216898612636
87 7.65850380086995
88 7.54918173293182
89 7.55047645670744
90 7.36051969031726
91 7.47198144857046
92 7.60092587053281
93 7.39182615096122
94 7.41630829930709
95 7.57552641167141
96 7.57669896958432
97 7.54566975950888
98 7.56776641434068
99 7.51541684625782
100 7.41198433684761
};
\addplot [semithick, color4]
table {%
36 13.9578908835618
37 13.0960013205241
38 10.8793122493475
39 10.2196959777161
40 10.1595828884015
41 9.59708976467062
42 9.61149595161109
43 9.27339884881078
44 9.33473881712325
45 8.72817308105672
46 8.96205415073887
47 8.87534072622726
48 8.84617016270483
49 8.53435028480414
50 8.42014236470109
51 8.89998097611958
52 8.5179221324872
53 8.61752660395289
54 8.55828309734554
55 7.99999482557785
56 8.13104470995787
57 8.24730880768545
58 8.03473398870227
59 8.0164665027027
60 8.19861094310813
61 8.12455955443284
62 7.81788019869205
63 8.09756887589367
64 7.90744476775635
65 7.98476860911914
66 7.79317406340663
67 7.82195147622163
68 7.75142854724788
69 7.77959143670172
70 7.87245490125298
71 8.03801943749303
72 7.78082665151232
73 7.78731842367941
74 7.70887263190021
75 7.82672816107206
76 7.53686608776358
77 7.49759408633263
78 7.50954103050327
79 7.69599423998546
80 7.74217998799255
81 7.75249415817999
82 7.6295224920925
83 7.71362938793418
84 7.5038410396799
85 7.41347349988702
86 7.40964100219353
87 7.77285122977249
88 7.4814457441724
89 7.49144737403041
90 7.26747707483966
91 7.55157411577193
92 7.38540813775474
93 7.3819525732177
94 7.38556823394662
95 7.43287233911862
96 7.67409727305421
97 7.37472165009827
98 7.42256410023905
99 7.46671828675482
100 7.36689432704003
};
\end{axis}

\end{tikzpicture}
  \caption {
  \rev{
    Conditioning number of the covariance matrix~$\matr{\Sigma_w}$ obtained with multiple
    learning algorithms. Intuitively, a lower matrix condition number for~$\matr{\Sigma}_w$ translates 
    into more robustness and numerical stability. The conditioning number is presented in 
    logarithmic scale. Note that the condition number stabilizes around 6 demonstrations
    for Maximum A Posteriori (MAP), whereas Maximum Likelihood (MLE) and the Least Squares method (\alexp{})
    with different regularization values~$\lambda$ requires around 50 demonstrations.
    In consequence, to avoid numerical problems the Prior distributions should be used
    unless a very large amount of data is available.
  }
  }
  \label{fig:promp:cond}
\end{figure}
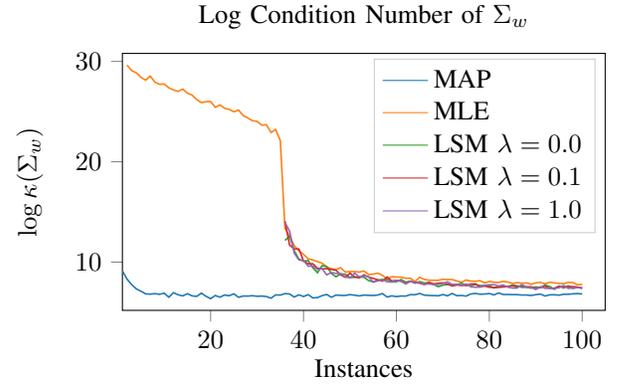

For~$\matr{\Sigma}_\omega$ we use an informative prior. 
Intuitively, the parameter~$v_0$ of the inverse Wishart prior represents 
how confident we are about our initial guess of the value of~$\matr{\Sigma}_\omega$ before 
looking at the data. We use~$v_0=\dim(\vect{\omega})+1$, that is the minimum value 
for~$v_0$ that results in a proper prior distribution~\cite{murphy2012machine}. We set the prior 
parameter~$\matr{S}_0$ as
\begin{equation}
  \matr{S}_0 = (v_0 + KD + 1)\blockdiag(\swmle),
  \label{eq:promp:s0}
\end{equation}
where~$\swmle$ is the maximum likelihood estimate of~$\matr{\Sigma}_w$ computed in 
line~\ref{alg:em_promp:mstep:cov_mle} of Algorithm~\ref{alg:em_promp}. 
Intuitively, the prior distribution favors considering joints independent when few 
data is available and gradually learn the correlation of different joints as more data 
is obtained. Using~\eqref{eq:promp:s0}, the update equation for~$\matr{\Sigma}_\omega$ 
on line~\ref{alg:em_promp:mstep:cov} of Algorithm~\ref{alg:em_promp} can be written
as
\[\matr{\Sigma}_\omega = \frac{1}{N + N_0}\left[N_0\blockdiag(\swmle) + N\swmle\right],\]
with~$N_0 = v_0 + KD + 1$. Note that the MAP estimate of~$\matr{\Sigma}_\omega$ is a linear
combination of the full MLE estimate and the MLE estimate under the assumption that all joints 
are independent. With a large number of trials~$N$, the MAP estimate will converge to the MLE
estimate as expected.
%

One of the reasons why we recommend using an informative prior for~$\matr{\Sigma}_\omega$, is because
its MLE estimate is typically numerically unstable.
We used the matrix condition 
number~$\kappa(\matr{\Sigma}_\omega)$ 
to measure numerical stability of the Maximum A Posteriori (MAP) and Maximum 
Likelihood Estimator (MLE) estimates of~$\matr{\Sigma}_\omega$. 
Intuitively, the condition number provides a measure of the sensitivity of an estimated value
to small changes in the input data~\cite{belsley2005regression}.
Therefore, a smaller the condition number means a more numerically stable estimate.
Figure~\ref{fig:promp:cond} shows the change in the condition number 
for~$\matr{\Sigma}_\omega$ in logarithmic scale with respect to the number of training
instances for both the MAP and MLE estimates. The condition number for the MAP estimate
is depicted in blue, and stabilizes around 6 training instances. On the other hand, the 
MLE estimate depicted in red requires around 50 training instances to stabilize.


\subsection{Relation to the Least Squares method to train ProMPs}
\label{sec:rel_alex_method}

\begin{algorithm}[t]
  \begin{algorithmic}[1]
    \Require Demonstration dataset containing the joint states and corresponding normalized time 
      stamps~$\matr{Y}=\{\vect{y}_{nt}, z_{nt}\}$ and the prior parameters $k_0, \vect{m}_0, v_0, \matr{S}_0$
    \Ensure The ProMP parameters~$\vect{\mu}_\omega$, $\matr{\Sigma}_\omega$, $\matr{\Sigma}_y$
    \State Compute matrices~$\matr{\phi}_{nt}=\matr{\phi}(z_{nt})$ with the basis functions~$\matr{\phi}$
    \State Compute $L = \sum_{n=1}^N{\tau_n}$
    \State Set some initial values for~$\vect{\mu}_\omega$, $\matr{\Sigma}_\omega$, $\matr{\Sigma}_y$. We
      use~$\vect{\mu}_\omega=\vect{0}$, $\matr{\Sigma}_\omega=\matr{I}$ and~$\matr{\Sigma}_y=\matr{I}$.
      \label{alg:em_approx:param_init}
    \While{Not converged}
      \For{ $n \in \{1,\dots,N\}$ }
        \State Compute~$\wpoint{n}$ with \eqref{eq:em_approx:point_est}
          \label{alg:em_approx:estep:point_est}
      \EndFor
      \State $\muwmle \gets \frac{1}{N}\left(\sum_{n=1}^N {\wpoint{n}}\right)$
        \label{alg:em_approx:mstep:mean_mle}
      \State $\vect{\mu}_\omega \gets \frac{1}{N + k_0}\left( k_0 \vect{m}_0 + N \muwmle \right)$
        \label{alg:em_approx:mstep:mean}
      \State $\swmle \gets \frac{1}{N}{\sum_{n=1}^N{\left((\wpoint{n} - \vect{\mu}_\omega)\trans{(\wpoint{n} - \vect{\mu}_\omega)}\right)}}$
        \label{alg:em_approx:mstep:cov_mle}
      \State $\matr{\Sigma}_\omega \gets \frac{1}{N + v_0 + KD + 1}\left[\matr{S}_0 + N\swmle\right]$
        \label{alg:em_approx:mstep:cov}
      \State $\matr{\Sigma}_y \gets \frac{1}{L}\sum_{n=1}^N{
        \sum_{t=1}^{\tau_n}{\left[
          \vect{\epsilon}_{nt}\trans{\vect{\epsilon}_{nt}}\right]}
      }$ \label{alg:em_approx:mstep:noise}
    \EndWhile
    \State \Return~$\vect{\mu}_\omega$, $\matr{\Sigma}_\omega$ and~$\matr{\Sigma}_y$.
  \end{algorithmic}
  \caption{EM training algorithm with a Dirac delta approximation for the E-step}
  \label{alg:em_approx}
\end{algorithm}

An alternative method of training ProMPs was proposed by~\cite{paraschos2015model}.
We show that the method proposed in~\cite{paraschos2015model} is a special case
of the EM algorithm presented in this paper for the MLE case, with a single iteration
and approximating the Gaussian distributions over the hidden variables~$\vect{\omega}_n$
with a Dirac delta distribution on the mean.

The method presented in~\cite{paraschos2015model} consists of making point estimates
of the hidden variables~$\vect{\omega}_n$ with least squares. Subsequently, the
mean and covariance of the point estimates are used to estimate the ProMP 
parameters. The point estimates of~$\vect{\omega}_n$ are computed for every trajectory 
using
\begin{equation}
  \vect{\omega}_n = (\trans{\matr{\Phi}_n}\matr{\Phi_n} + \lambda\matr{I})^{-1}
  \trans{\matr{\Phi}_n}\vect{y}_n,
  \label{eq:point_est:w}
\end{equation}
where~$\matr{\Phi}_n$ and~$\vect{y}_n$ are the vertical concatenation of the 
matrices~$\matr{\Phi}_{nt}$ and vectors~$\vect{y}_{nt}$ respectively, and $\lambda$ is 
a ridge regression parameter that can be set to zero unless numerical
problems arise. Subsequently, the ProMP parameters can be estimated using the MLE
estimates for Gaussian distributions
\begin{align}
  \muwmle    &= \frac{1}{N}\sum\nolimits_{n=0}^N{\vect{\omega}_n},
  \label{eq:point_est:mle:mu_w} \\
  \swmle &= \frac{1}{N}{ 
    \sum\nolimits_{n=0}^N {
      (\vect{\omega}_n - \vect{\mu}_\omega)
      \trans{(\vect{\omega}_n - \vect{\mu}_\omega)}
    }.
  }
  \label{eq:point_est:mle:sigma_w}
\end{align}

\rev{
Figure~\ref{fig:promp:cond} shows the numerical stability of the 
matrix~$\matr{\Sigma}_\omega$ of equations~\eqref{eq:point_est:w} 
to~\eqref{eq:point_est:mle:sigma_w} using multiple values of~$\lambda$.
Note that this training procedure has the same numerical stability issues that the MLE
estimates independently of the value of $\lambda$. The reason is that the numerical
problems do not come from the estimation of~$\omega$, where~$\lambda$ is being used,
but from the estimation of the covariance matrix itself on~\eqref{eq:point_est:mle:sigma_w}.
Note also that Figure~\ref{fig:promp:cond} displays the conditioning number 
of~$\matr{\Sigma}_\omega$ for this method using a minimum of 36 demonstrations. The reason
is that using~\eqref{eq:point_est:mle:sigma_w} with~$N < KD$ would result in a rank
deficient matrix~$\matr{\Sigma}_\omega$ whose condition number would be~$+\infty$.
}

\rev{
  The discussed numerical issues of training a ProMP using~\eqref{eq:point_est:w} 
  to~\eqref{eq:point_est:mle:sigma_w} make the learned model dangerous to use directly 
  for robotic applications. In~\cite{paraschos2015model}, a small diagonal matrix is
  added to~$\matr{\Sigma}_\omega$ and in addition a artificial noise matrix~$\matr{\Sigma}_y^*$
  needs to be used during conditioning. A very large number of training instances~$N >> KD$ would
  be required to be able to use~\eqref{eq:point_est:w} to~\eqref{eq:point_est:mle:sigma_w}
  without any additional tricks. Using the proposed prior distribution solves the numerical
  problems in a theoretically sound way, and in the limit of infinite amount of data it
  converges to the expected estimation procedure using maximum likelihood.
}

\rev{However, the estimation method proposed in Algorithm~\ref{alg:em_promp} differs
from using~\eqref{eq:point_est:w} to~\eqref{eq:point_est:mle:sigma_w} in more than
just using a prior distribution. To show the differences and similarities,
} let us now analyze the 
EM algorithm presented in this paper if we approximate the E-step
with a Dirac delta distribution. Note that using a Dirac delta 
distribution~$\delta(\vect{\omega}-\wpoint{n})$ means making a point estimate~$\wpoint{n}$
of the hidden variables~$\vect{\omega}_n$ without any uncertainty.
The value of the point estimates~$\wpoint{n}$ is given by
\begin{equation}
  \resizebox{0.85\hsize}{!}{
  $\wpoint{n} = \left(\matr{\Sigma}_\omega^{-1} + 
          \trans{\matr{\Phi}_{n}}\matr{\Sigma}_y^{-1}\matr{\Phi}_{t}\right)^{-1}
          \left(\matr{\Sigma}_\omega^{-1}\vect{\mu}_\omega +
          \trans{\matr{\Phi}_{n}}\matr{\Sigma}_y^{-1}\vect{y}_{n}\right).$
        }
  \label{eq:em_approx:point_est}
\end{equation}
Algorithm~\ref{alg:em_approx} shows the resulting EM algorithm with the discussed approximation
for the E-step. The quality of the approximation depends on how much uncertainty
is there in the computation of the hidden variables. Let us further assume that we execute one
single iteration of Algorithm~\ref{alg:em_approx} with initial 
values~$\matr{\Sigma}_\omega = \lambda^{-1}\matr{I}$,
$\vect{\mu}_\omega = 0$ and~$\matr{\Sigma}_y = \matr{I}$. It is easy to see that the
estimates~$\wpoint{n}$ would be exactly equivalent to the estimates~\eqref{eq:point_est:w} used 
by~\cite{paraschos2015model}. Note also that Lines~\ref{alg:em_approx:mstep:mean_mle} 
and~\ref{alg:em_approx:mstep:cov_mle} of Algorithm~\ref{alg:em_approx} compute also exactly
the same estimates of~\cite{paraschos2015model} on~\eqref{eq:point_est:mle:mu_w}
and~\eqref{eq:point_est:mle:sigma_w} for the ProMP parameters in the MLE case.

We can conclude that the training procedure from~\cite{paraschos2015model} is equivalent
to a single iteration of the approximated training procedure presented in 
Algorithm~\ref{alg:em_approx} on the MLE case with a particular initialization of 
the ProMP parameters. 
\rev{
We have already extensively discussed the advantages of using MAP estimates using the proposed
prior distribution. The remaining questions we want to discuss are weather using 
uncertainty estimates and more than one EM iteration is helpful.
}

\rev{Estimating the uncertainty helps in applications where there is actually high uncertainty
in the estimation of the hidden variables~$\vect{\omega}_n$ due for example to missing
observations or high sensor noise.
The answer of how much multiple iterations help depends entirely on the parameter initialization
(see Line~\ref{alg:em_approx:param_init} of Algorithm~\ref{alg:em_approx}). Note that the
only difference between the first iteration and the rest is that in the first iteration we
are working entirely on our initial guess of the values of the ProMP parameters. Whereas
in subsequent iterations we are using optimized estimates of the ProMP parameters.
}

\rev{
For our robot experiments, the sensor noise is on the order of~$10^{-3}$ radians and the number
of samples per trajectory is between 200 and 500 per degree of freedom. Furthermore,
there are no missing observations as we can always read the joint sensor values.
With such a low signal to noise ratio and without missing observations the values of the hidden
variables~$\vect{\omega}_n$ can be estimated very precisely and with low uncertainty.
As we expected, we did not observe any difference in the performance in any our our robot 
experiments using the estimates produced by Algorithms~\ref{alg:em_promp} 
and~\ref{alg:em_approx}. In fact, the estimated parameters~$\vect{\mu}_\omega$ 
and~$\matr{\Sigma}_\omega$ produced by both algorithms are virtually the same.
We can conclude that for modeling robot trajectories on a robot setup like ours, 
using the approximation of the E-step with a Dirac delta distribution does not 
impact the performance compared to the complete EM estimation.
}

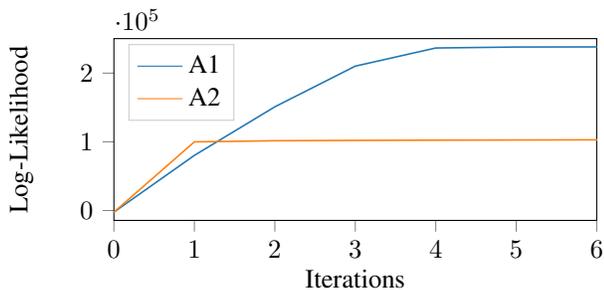
\begin{figure}
  \centering
  \setlength\figurewidth{8cm}
  \setlength\figureheight{4cm}
  \setlength\axislabelsep{0mm}
  \tikzsetnextfilename{emballlhood}
\begin{tikzpicture}

\definecolor{color1}{rgb}{1,0.498039215686275,0.0549019607843137}
\definecolor{color0}{rgb}{0.12156862745098,0.466666666666667,0.705882352941177}

\begin{axis}[
xlabel={Iterations},
ylabel={Log-Likelihood},
xmin=0.0, xmax=6,
ymin=-14358.3580216542, ymax=250186.763611088,
width=\figurewidth,
height=\figureheight,
tick align=outside,
tick pos=left,
x grid style={lightgray!92.026143790849673!black},
y grid style={lightgray!92.026143790849673!black},
legend cell align={left},
legend entries={{A1},{A2}},
legend style={at={(0.03,0.97)}, anchor=north west, draw=white!80.0!black}
]
\addlegendimage{no markers, color0}
\addlegendimage{no markers, color1}
\addplot [semithick, color0]
table {%
0 -2333.46356316458
1 80251.0467293008
2 150941.130394339
3 210137.192515681
4 236542.553317696
5 237956.133367132
6 238089.195924563
7 238161.985355055
};
\addplot [semithick, color1]
table {%
0 -2333.57976562046
1 99917.6424352217
2 101567.951322142
3 102084.388153255
4 102419.621091013
5 102681.76921478
6 102895.111359865
7 103071.159536897
};
\end{axis}

\end{tikzpicture}
  \caption{
    Log-Likelihood improvement with every iteration of the proposed EM algorithm presented
    in Algorithm~\ref{alg:em_promp} (A1) and the approximated version presented in 
    Algorithm~\ref{alg:em_approx} (A5). Both algorithms make initially
    poor estimates of the hidden variables~$\vect{\omega}_n$. However, the
    proposed algorithm also captures the uncertainty over the estimates of
    the hidden variables, opposed to the approximated algorithm. As a result,
    the proposed algorithm continues to successfully improve the likelihood
    after the first iteration, whereas the approximated algorithm improves only
    marginally after the first iteration.
  }
  \label{fig:point_est:em_improve}
\end{figure}

\rev{
  To show an example problem where uncertainty estimates are more important, 
  we decided to run a small additional experiment where we use a ProMPs to model 
  a table tennis ball trajectory. We collected 80 ball trajectories using the
  robot vision system, there are a few missing observations due to occlusion or
  errors in the image processing algorithms as well as a higher signal to noise 
  ratio. Subsequently, we trained two models
  using the exact and approximated training algorithms. Finally, we tested 
  the trained models predicting the ball position at time~$t=1.2 s$ given the
  first 160 milliseconds of ball observations. On this experiment, we used~$K=8$
  basis functions and~$D=3$ dimensions. The initial parameters for both algorithms
  were~$\matr{\Sigma}_\omega = \matr{I}$, $\vect{\mu}_\omega = 0$ 
  and~$\matr{\Sigma}_y = \matr{I}$
}

Figure~\ref{fig:point_est:em_improve} shows the evaluation
of the log likelihood for each iteration of the EM algorithm for both the proposed
and the approximated versions. Both algorithms are provided the exact same
data and use the exact same parameter initialization. Note that the proposed
algorithm outperforms the approximated algorithm in this particular problem.
\rev{
  The distance between the ground truth ball position measured by the vision system and
  the position predicted by the ProMP models trained with Algorithm~\ref{alg:em_promp} is 
  around~\textbf{10 cm}, 
  whereas the error of the ProMP trained with Algorithm~\ref{alg:em_approx} is 
  around~\textbf{50 cm}.
}

As a final argument in favour or using Algorithm~\ref{alg:em_promp} instead of
Algorithm~\ref{alg:em_approx}, note that we are not gaining anything out of the
approximation. The algorithmic complexity is exactly the same in both cases, and
estimating the uncertainty does not hurt the learning algorithm even on the cases
where it is very low and the approximation seems to be accurate.

\section{Adaptation of Probabilistic Movement Primitives}
\label{sec:promp:operators}

\begin{figure*}
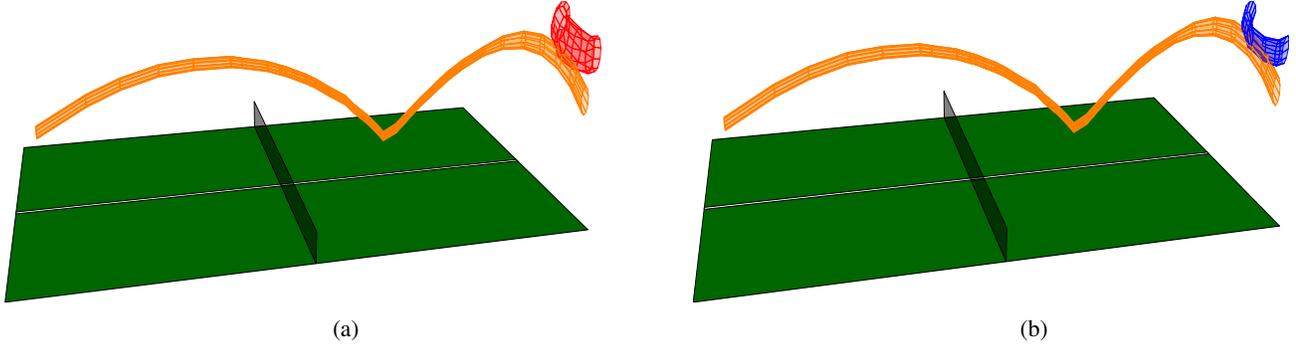

  \begin{subfigure}[b]{0.5\textwidth}
    \tikzsetnextfilename{tableprior}
    \resizebox{8cm}{4cm}{%
      \input{fig/task_dist/table_prior.tex}
    }
    \caption{}
    \label{fig:table_dist:prior}
  \end{subfigure}
  \begin{subfigure}[b]{0.5\textwidth}
    \tikzsetnextfilename{tablepost}
    \resizebox{8cm}{4cm}{%
      \input{fig/task_dist/table_post.tex}
    }
    \caption{}
    \label{fig:table_dist:post}
  \end{subfigure}
  \caption{
    Task space distributions of the ball and the racket center before and after
    adapting the ProMP in task space. The distribution of the ball is presented in orange.
    Figure~\ref{fig:table_dist:prior} depicts in red the distribution of the center of the 
    racket computed from the ProMP learned from human demonstrations. Subsequently, the ProMP is adapted
    to hit the ball using Algorithm~\ref{alg:mod_end_eff}, and the resulting ProMP is
    depicted in blue in Figure~\ref{fig:table_dist:post}. Note that the adapted ProMP is
    similar to the original ProMP learned from human demonstrations, but the probability
    mass is concentrated in the area that overlaps with the ball trajectory distribution.
  }
  \label{fig:table_dist}
\end{figure*}

Adapting a movement primitive by setting initial positions, desired via points or
final positions is a necessary property to generalize to different situations. These
desired via points could be specified in joint space or in task space. For example,
a table tennis striking movement needs to start in the current joint configuration
of the robot and later reach the predicted task space position of the table tennis
ball. In this section, we present operators to adapt movement primitives in joint
and in task space. In addition, we evaluate the execution time of these operators
showing that they can all run in less than one millisecond on a standard computer,
satisfying the real time requirements of the applications presented in this paper.

\subsection{Adapting a ProMP in Joint Space}

In the original formulation of
ProMPs~\cite{proMP}, it was proposed to adapt a ProMP in joint space by conditioning
on a desired observation~$\ytdes$ with some noise matrix~$\matr{\Sigma}_y^{*}$ that
was referred to as the desired accuracy. However, the authors do not provide any
intuition on how~$\matr{\Sigma}_y^{*}$ should be computed or estimated.

In this paper, we follow the approach presented in previous work~\cite{gomez2016using} to 
adapt in joint space by conditioning in the distribution over
joint trajectories to reach a particular value~$\vect{y}_t=\ytdes$ without any artificial
accuracy matrix~$\matr{\Sigma}_y^{*}$. The reason why we do not need the artificial noise
matrix~$\matr{\Sigma}_y^{*}$ opposed to~\cite{proMP} is that we do not suffer from
numerical problems inverting~$\matr{\Sigma}_\omega$ due to the different training procedure.
The conditioned
distribution~$p(\vect{\omega}|\vect{y}_t = \ytdes) \propto p(\vect{y}_t = \ytdes|\vect{\omega})p(\vect{\omega})$
can be computed in closed form and is given by
\begin{align*}
  p(\vect{\omega}|\vect{y}_t = \ytdes) &= \normal{\vect{\omega}}{\vect{m}_\omega}{\matr{S}_\omega}, 
  \\
  \vect{m}_\omega &= \matr{S}_\omega(\trans{\matr{\phi}_t}\matr{\Sigma}_y^{-1}\ytdes + \matr{\Sigma}_\omega^{-1}\vect{\mu}_\omega),
  \\
  \matr{S}_\omega &= (\matr{\Sigma}_\omega^{-1} + \trans{\matr{\phi}_t}\matr{\Sigma}_y^{-1}\matr{\phi}_t)^{-1}.
\end{align*}
There are cases where we do not know the exact value of the desired joint 
configuration~$\ytdes$, but instead we have a probability 
distribution~$\ytdes \sim \normal{\ytdes}{\vect{\mu}_q}{\matr{\Sigma}_q}$.
For example, in the table tennis task, we need condition the striking movement 
on the future position of the ball predicted using a Kalman filter. The distribution
of the ball is, however, in task space. In Section~\ref{sec:promp:invkin}, we
explain how to transform a target task space distribution to a joint space
distribution. For the moment, we assume we have a target distribution in
joint space, that we can marginalize using
\[p(\vect{\omega}|\vect{\mu}_q,\matr{\Sigma}_q) = \int{p(\vect{\omega}|\vect{y}_t = \ytdes)\normal{\ytdes}{\vect{\mu}_q}{\matr{\Sigma}_q}\diff{\ytdes}},\]
which can be computed in closed form obtaining
\begin{align}
  p(\vect{\omega}|\vect{\mu}_q,\matr{\Sigma}_q) &= \normal{\vect{\omega}}{\vect{m}_\omega}{\matr{S}_\omega},
  \nonumber \\
  \vect{m}_\omega &= \matr{S}_\omega(\trans{\matr{\phi}_t}\matr{\Sigma}_y^{-1}\vect{\mu}_q + \matr{\Sigma}_\omega^{-1}\vect{\mu}_\omega),
  \label{eq:promp:cond:dist:mean_raw} \\
  \matr{S}_\omega &= \matr{T}_\omega + \matr{T}_\omega\trans{\matr{\phi}_t}\matr{\Sigma}_y^{-1}\matr{\Sigma}_q
  \matr{\Sigma}_y^{-1}\matr{\phi}_t\matr{T}_\omega,
  \label{eq:promp:cond:dist:cov_raw}
\end{align}
with~$\matr{T}_\omega = (\matr{\Sigma}_\omega^{-1} + \trans{\matr{\phi}_t}\matr{\Sigma}_y^{-1}\matr{\phi}_t)^{-1}$.
The ProMPs can also be adapted with desired velocities or accelerations using the same method,
replacing the basis function matrices~$\matr{\phi}_t$ by their respective time derivatives~$\dot{\matr{\phi}_t}$
and~$\ddot{\matr{\phi}_t}$.

The run time complexity for both adaptation operators is bounded by~$O(K^{3}D^{3})$. In our
robot experiments we used a model with~$KD=35$, obtaining an average execution time of 0.044~ms.
In the experimental section, we provide running times for different model sizes.

The methods presented here to condition in a ProMP on a particular joint configuration~$\ytdes$
and desired joint distribution were already introduced in previous work~\cite{gomez2016using}.
The comparison with the adaptation operator presented in~\cite{proMP} and the analysis of their
execution time is new to this paper.

\subsection{Probability Distribution of a ProMP in Task Space}
\label{sec:promp:kinematics}

We compute a probability distribution in task space from a ProMP
learned in joint space making use of the
geometry of the robot. We assume that we have access to a deterministic 
function~$\vect{x}_t = \vect{f}(\vect{y}_t)$ called the forward kinematics function that 
returns the position in task space $\vect{x}_t$ of a point of interest like the end effector
of the robot given the joint state configuration $\vect{y}_t$. The deterministic forward
kinematics function~$\vect{f}$, can be expressed in our probabilistic framework using
\begin{equation*}
  p(\vect{x}_t | \vect{y}_t) = \delta(\vect{x}_t - \vect{f}(\vect{y}_t)),
  \label{eq:prob_fwd_kin}
\end{equation*}
where $\delta$ is the Dirac delta function. The task space distribution can be computed from 
the ProMP parameters learned in joint space using
\[ p(\vect{x}_t|\vect{\theta}_\omega) = \int{
  p(\vect{y}_t|\vect{\theta}_\omega) p(\vect{x}_t | \vect{y}_t) d\vect{y}_t}.
\]
The distribution~$p(\vect{x}_t|\vect{\theta}_\omega)$ can not be computed in
closed form for a non-linear forward kinematics function. We compute an
approximated distribution~$p(\vect{x}_t|\vect{\theta}_\omega)$ making a linear Taylor
expansion of the forward kinematics function around the ProMP mean, obtaining
\begin{equation}
  p(\vect{x}_t|\vect{\theta}_\omega) = \normal{
    \vect{x}_t
  }{
    \vect{f}(\matr{\Phi}_t\vect{\mu}_\omega)
  }{
    \matr{J}_t\matr{\Sigma}_\omega\trans{\matr{J}_t}
  },
  \label{eq:promp:task_dist} 
\end{equation}
where~$\matr{J}_t=\matr{J}(\matr{\Phi}_t\vect{\mu}_\omega)$ is the Jacobian of the forward
kinematics function~\cite{siciliano2010robotics} evaluated 
at~$\vect{y}_t = \matr{\Phi}_t\vect{\mu}_\omega$.
Figure~\ref{fig:table_dist:prior} shows the task space distribution of a ProMP learned
from demonstrations to strike a table tennis ball as well as some particular ball 
trajectory distribution. The distribution of the center racket is depicted in red
and the distribution of the predicted ball trajectory is depicted in orange. 

%
\subsection{Adapting ProMPs in Task Space}
\label{sec:promp:invkin}

For many applications, it is more natural to define goals in task space. For instance,
in robot table tennis the movement primitive should be adapted such that the position of
the racket matches the predicted position of the ball. In this section, we present our 
approach to condition a ProMP learned in joint space to have a desired task space 
distribution. The approach we present in this section was also introduced in
previous work~\cite{gomez2016using}. However, in this paper we evaluate it more thoroughly
on a coffee preparation task with clear training and validation set to assess for
generalization.

We denote the desired task space state at time $t$ by the random 
variable $\vect{x}_t$, with probability distribution given by
\begin{equation*}
  p(\vect{x}_t|\vect{\theta}_x) = \normal{\vect{x}_t}{\vect{\mu}_x}{\matr{\Sigma}_x},
  \label{eq:promp:inv_kin:prior_x}
\end{equation*}
where the parameters~$\vect{\theta}_x = \{\vect{\mu}_x, \matr{\Sigma}_x\}$ are user
inputs that represent the desired task space configuration and its uncertainty 
respectively. 
For the table tennis task, we use a Kalman filter as ball trajectory model. The Kalman
filter provides an estimate for the mean ball position~$\vect{\mu}_x$ and its
uncertainty~$\matr{\Sigma}_x$.

Given a desired end effector position~$\vect{x}_t$ and a ProMP with 
parameters~$\vect{\theta}_\omega=\{\vect{\mu}_\omega,\matr{\Sigma}_\omega\}$, 
a probability distribution for the joint configuration 
can be computed by 
\begin{equation}
  p(\vect{y}_t|\vect{x}_t,\vect{\theta}_\omega) \propto p(\vect{x}_t|\vect{y}_t)p(\vect{y}_t|\vect{\theta}_\omega),
  \label{eq:promp:inv_kin:point_post}
\end{equation}
where~$p(\vect{x}_t|\vect{y}_t)$ is given by~\eqref{eq:prob_fwd_kin} 
and~$p(\vect{y}_t|\vect{\theta}_\omega)$ is the joint space distribution given by 
the ProMP.

The distribution~$p(\vect{y}_t|\vect{x}_t,\vect{\theta}_\omega)$ represents a 
compromise between staying close to the demonstrated trajectories and achieving the 
desired racket configuration. Thus, for a robot arm with redundant degrees of freedom, 
where multiple joint space configurations can achieve the desired racket configuration, 
the presented approach will prefer joint solutions that are closer to the 
demonstrated behavior.

To achieve the desired task space distribution~$p(\vect{x}_t|\vect{\theta}_x)$ instead 
of a particular value~$\vect{x}_t$, we marginalize out~$\vect{x}_t$ 
from~\eqref{eq:promp:inv_kin:point_post} obtaining
\begin{equation}
  \begin{split}
    p(\vect{y}_t|\vect{\theta}_x,\vect{\theta}_\omega) &= \int{p(\vect{y}_t|\vect{x}_t,\vect{\theta}_\omega)p(\vect{x}_t|\vect{\theta}_x)\diff{\vect{x}_t}} \\
                                  &\propto  p(\vect{y}_t|\vect{\theta}_\omega)\int{p(\vect{x}_t|\vect{\theta}_x)p(\vect{x}_t|\vect{y}_t)\diff{\vect{x}_t}}.
  \end{split}
  \label{eq:promp:inv_kin:post_raw}
\end{equation}
Note that~$p(\vect{y}_t|\vect{\theta}_x,\vect{\theta}_\omega)$ is a distribution in joint
space that again compromises between staying close to the demonstrated behavior and achieving
the desired task space distribution. The integral in~\eqref{eq:promp:inv_kin:post_raw}, required 
to compute the normalization constant of~$p(\vect{y}_t|\vect{\theta}_x,\vect{\theta}_\omega)$, is
intractable. We used Laplace Approximation~\cite{bishop2006pattern} to compute a Gaussian 
approximation for~$p(\vect{y}_t|\vect{\theta}_x,\vect{\theta}_\omega)$. With a Gaussian 
distribution for~$p(\vect{y}_t|\vect{\theta}_x,\vect{\theta}_\omega)$, the operator
to adapt ProMPs in joint space discussed in Section~\ref{sec:promp:operators} can be
used to obtain a new adapted ProMP.

\begin{algorithm}[t]
  \begin{algorithmic}[1]
    \Require Parameters of desired task space distribution~$\vect{\theta}_x=[\vect{\mu}_x, \matr{\Sigma}_x]$ 
    and ProMP to adapt~$\vect{\theta}_\omega=[\vect{\mu}_\omega, \matr{\Sigma}_\omega]$.
    \Ensure A new ProMP modulated to strike the ball
    \State $\vect{\mu}_q \gets \argmax_{\vect{y}_t}{\left(\log{p(\vect{y}_t|\vect{\theta}_x,\vect{\theta}_\omega)}\right)}$
      \label{alg:mod_end_eff:mean}
    \State Compute $\matr{\Lambda}_q$ as the second derivative of $\log{p(\vect{y}_t|\vect{\theta}_x, \vect{\theta}_\omega)}$
      with respect to $\vect{y}_t$ evaluated at $\vect{y}_t = \vect{\mu}_q$
      \label{alg:mod_end_eff:hessian}
    \State $\matr{\Sigma}_q \gets \matr{\Lambda}_q^{-1}$
      \label{alg:mod_end_eff:cov}
    \State Compute $\vect{m}_\omega$ and $\matr{S}_\omega$ with \eqref{eq:promp:cond:dist:mean_raw} and
      \eqref{eq:promp:cond:dist:cov_raw}    
    \State \Return new ProMP with $\vect{\mu}_\omega=\vect{m}_\omega$ and $\matr{\Sigma}_\omega=\matr{S}_\omega$
  \end{algorithmic}
  \caption{Algorithm to adapt a ProMP in task space using Laplace Approximation}
  \label{alg:mod_end_eff}
\end{algorithm}

Algorithm~\ref{alg:mod_end_eff} describes the procedure to adapt a ProMP in task space using
Laplace Approximation. The mean~$\vect{\mu}_q$ and covariance~$\matr{\Sigma}_q$ of the 
approximated joint space distribution are computed in Lines~\ref{alg:mod_end_eff:mean} 
and~\ref{alg:mod_end_eff:cov} respectively. The presented operator for task space conditioning
consists of a non linear optimization to compute~$p(\vect{y}_t|\vect{\theta}_x,\vect{\theta}_\omega)$
followed by a use of the joint space conditioning operator. As a result, the task space 
conditioning operator is necessarily slower than the joint space conditioning operator.
In Section~\ref{sec:promp:operators:runtime}, we show that the execution time of the
presented operator is nonetheless reliably below 3 milliseconds for ProMP sizes up 
to~$KD=350$, satisfying the real time requirements of our robot applications by a
large margin.

Figure~\ref{fig:table_dist} depicts the task space distribution of a ProMP
learned from forehand strike demonstrations before~\ref{fig:table_dist:prior} 
and after~\ref{fig:table_dist:post} adapting it to hit a ball trajectory seen
at test time. Note that the adapted ProMP has the probability mass concentrated
in the region that overlaps with the ball trajectory distribution.

\subsection{Execution Time of the Presented Operators}
\label{sec:promp:operators:runtime}

\begin{table}
  \centering
  \begin{tabular}{@{}lll@{}}
    \toprule
    {\bf $KD$} & {\bf Joint Space [ms]} & {\bf Task Space [ms]} \\
    \midrule
    35 & $0.0448 \pm 0.0164$ & $0.7212 \pm 0.2920$ \\
    70 & $0.0642 \pm 0.0104$ & $0.8328 \pm 0.5484$ \\
    140 & $0.1880 \pm 0.0245$ & $1.0764 \pm 0.3179$ \\
    210 & $0.5294 \pm 0.5879$ & $1.4291 \pm 0.2423$ \\
    280 & $0.8686 \pm 0.7944$ & $1.9267 \pm 0.3822$ \\
    350 & $1.2095 \pm 0.4829$ & $2.3173 \pm 0.3135$ \\
    \bottomrule
  \end{tabular}
  \caption{
    Average execution time of joint and task space conditioning operators 
    in milliseconds for ProMPs of different sizes. The task space operator
    uses internally the joint space operator, as a result is has a higher execution time.
    The size of a ProMP is given by the product between the number of degrees
    of freedom~$D$ and the number of kernels per degree of freedom~$K$.
    The table presents the mean and standard deviation of the running times 
    for each operator in milliseconds.
    For robot table tennis, all these operators need to be executed after the
    ball trajectory is predicted using ball observations and a ball model.
    In consequence, it is crucial to be able to apply these operators fast enough
    to successfully hit the already flying ball.
  }
  \label{tab:results:time}
\end{table}

Many use cases for the operators presented in this paper to adapt the movement 
primitives will have real time execution requirements. If we want to adapt a
movement primitive with respect to sensor values measured at time~$t_1$ and 
subsequently execute the movement primitive in the robot at time~$t_2$, the
total execution time for the operator cannot exceed $t_2-t_1$.

For example, to make sure that the executed movement primitive starts on the current
robot joint state, we use the joint conditioning operator on the measured joint state
just before starting the execution of the movement primitive. For our robot experiments,
we used a control loop of 500 Hertz. Therefore, we have a real time constrain of 2~ms
to read the sensor value for the joint state, condition the ProMP to start at the measured
value and send the required motor commands. 

Table~\ref{tab:results:time} show the average execution time and standard deviation in 
milliseconds for the operators presented in this paper. Each operator
is executed 1000 times for each of the different sizes of ProMPs in 
a Lenovo Thinkpad X2 Carbon laptop with a processor
Intel Core i7-6500U 2.50GHz and 8 GB of RAM. We report the size as the
product of the degrees of freedom~$D$ and the number of basis functions per
degree of freedom~$K$.

On our robot experiment we used a ProMP with size~$K=5$ and~$D=7$, that
corresponds to the smallest entry in Table~\ref{tab:results:time}.
However, note that even on a ProMP with~$KD=350$ we can meet the real time requirements 
to play robot table tennis. The operator to condition in joint space can be
reliably run under the 2 milliseconds required for our control loop of 500 Hertz. 
The vision system we use in this paper produces 60 ball observations per second. 
Therefore, we can potentially correct the ProMP trajectory to changes in the ball
trajectory after every ball observation with a running time below 16 milliseconds.
Note that our task space conditioning operator runs reliably under 3 milliseconds,
satisfying the real time requirements by a large margin.

\section{Experiments and Results}
\label{sec:experiments}

\begin{figure}
  \centering
  \setlength\figurewidth{8cm}
  \setlength\figureheight{4cm}
  \setlength\axislabelsep{0mm}
  \tikzsetnextfilename{superposed}
  \input{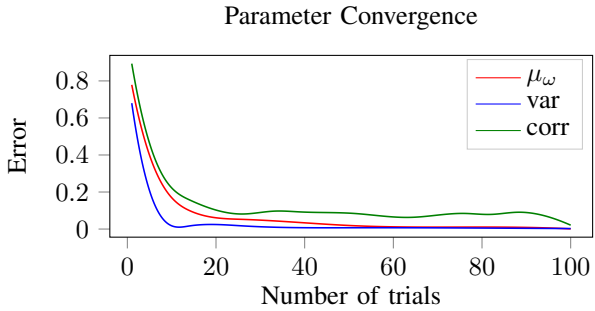}
  \caption{
    Convergence of the ProMP parameters as a function of the number of training 
    instances in an adversarial scenario. The convergence of different sets of 
    parameters is depicted with
    different colors. The set of parameters corresponding to the mean behavior,
    variability of the movement, and correlation between joints are depicted in 
    red, blue and green respectively.
  }
  \label{fig:conv:superposed}
\end{figure}

We evaluate the presented methods with synthetic data and
with a real robot experiments for table tennis and assisting coffee brewing.
For the robot experiments we used Barrett WAM arm with seven degrees of freedom
capable of high speed motion. The robot control computer uses a 500~Hz control
loop, receiving joint angle measures and output motor commands every 2~ms.
To track the position of objects of interest like the table tennis ball and
the coffee machine, we used four Prosilica Gigabit cameras and the vision system
described in~\cite{lampert2012vision}. This vision system tracks the position
of a table tennis ball with an approximate frequency of 60~Hz, we attached a
table tennis ball to the coffee machine for the coffee brewing experiments. 

On all our robot experiments we used five basis functions per degree of
freedom. Fifth~\cite{mulling2011biomimetic} and third~\cite{KOC2018} order 
polynomials have been previously used successfully for robot
table tennis approaches. Note that the same results should be achievable with a 
ProMP with six or four polynomial basis functions respectively taking into
account the constant term.
On the other hand, radial basis functions (RBFs) have been typically used with ProMPs~\cite{proMP}
for other robot applications. We tried different combinations of RBFs 
and polynomial basis functions, obtaining the best results using three RBFs
and a first order polynomial, for a total of five basis functions.

\subsection{Parameter Convergence on Synthetic Data}

The purpose of the experiment with synthetic data, is to evaluate 
how accurate are the estimates of the ProMP
parameters as a function of the number of training instances~$n$ when
the assumptions we made for the prior distribution are incorrect. 
We generate synthetic data from a reference ProMP that displays a
strong correlation between different degrees of freedom, opposing the
proposed prior assumptions. Subsequently, we test
if the proposed learning procedure converges to the expected
parameters and how many training examples are necessary for
convergence.

\begin{figure}
  \centering
  \includegraphics[width=6cm]{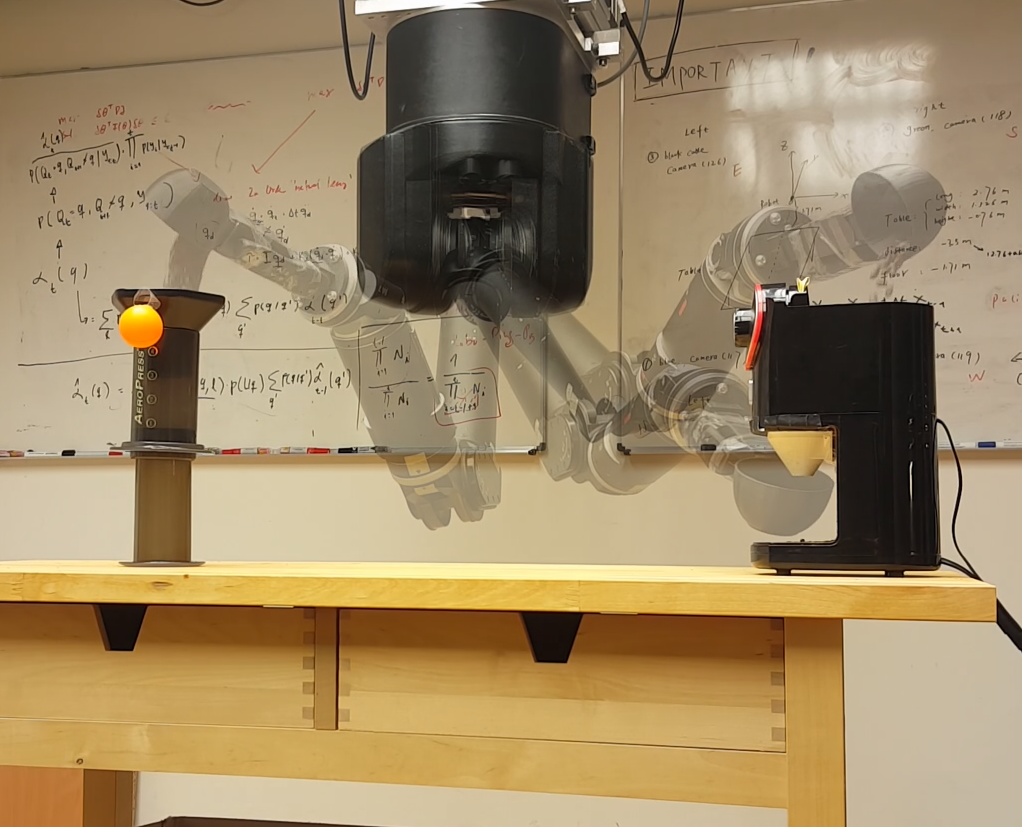}
  \caption{
    The robot executing the coffee task. First, the robot moves towards the top
    of the coffee grinder to pour fresh beans into it. Subsequently, the robot moves towards 
    the bottom of the grinder to pick the grounds. Finally, the robot deposits the
    coffee grounds in the brewing chamber of the coffee machine.
  }
  \label{fig:coffee}
\end{figure}

On this synthetic data experiment there is no notion of training
or test sets. We simply generate $n$~sample trajectories from a reference 
ProMP with known parameters~$\vect{\mu}_\omega$ and~$\matr{\Sigma}_\omega$.
Subsequently, we train a new ProMP with the sampled trajectories obtaining a 
new set of parameters~$\hat{\vect{\mu}}_\omega^n$ and~$\hat{\matr{\Sigma}}_\omega^n$
and compare how close they are to the reference parameters~$\vect{\mu}_\omega$ 
and~$\matr{\Sigma}_\omega$ using the Frobenius norm. 
In this experiment we used five basis functions~$K=5$ and
four degrees of freedom~$D=4$. 
To ensure a high correlation, we set the parameters
of the base ProMP such that the last two degrees of freedom are the addition
and subtraction of the first two degrees of freedom respectively.


Figure~\ref{fig:conv:superposed} show the average parameter estimation error 
with respect to the number of training instances~$n$ for different set of
parameters. The error over the parameters~$\vect{\mu}_\omega$ that represent the
mean behavior is depicted in red. The error over the parameters~$\matr{\Sigma}_\omega$
are divided in the block diagonal terms that represent the captured variability
of the movement (depicted in blue) and the rest of the parameters that represent 
the captured joint correlations (depicted in green). The error of the different
set of parameters is normalized between zero and one to facilitate comparison,
and the error curves are smoothed out using splines to facilitate visualization of
convergence. 


Note that the learning algorithm converges to the true value as expected. However,
more training examples are required to converge to the correlation parameters because
the prior is favouring joint independence in a high joint correlation scenario.
The effect of the proposed prior is to prefer 
independence between the joints in absence of strong evidence of correlation. 


The results from this experiment may suggest that the presented probabilistic 
framework require large amounts of data samples to learn a movement primitive.
In contrast, we show we can learn a coffee-pouring and a table-tennis 
experiment that the proposed approach, using only two and eight training
examples respectively. There are two main explanations why we can converge with
fewer training instances to the target performance on different tasks.
First, the prior distribution assumptions may be more accurate in some real
world tasks than in the adversarial example chosen in this section. Second, 
we can not compare convergence in parameter space to convergence in the
performance of a particular task. The reason is that there might be multiple
different parameter values with a similar task performance.

\subsection{Assisting Coffee Brewing}

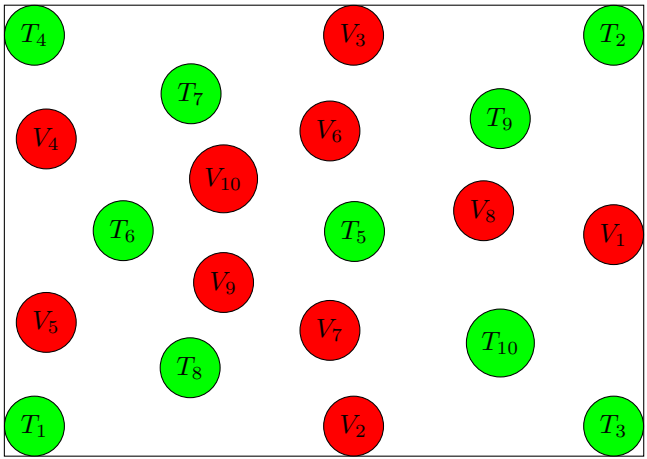
\begin{figure}
  \centering
  \tikzsetnextfilename{coffeepattern}
  \begin{tikzpicture}
    \draw (0cm,0cm) rectangle (8.5cm,6.0cm);
    \node[draw,circle,fill=green] at (0.4cm,0.4cm) {$T_1$};
    \node[draw,circle,fill=green] at (8.1cm,5.6cm) {$T_2$};
    \node[draw,circle,fill=green] at (8.1cm,0.4cm) {$T_3$};
    \node[draw,circle,fill=green] at (0.4cm,5.6cm) {$T_4$};
    \node[draw,circle,fill=green] at (4.65504201681cm,2.98893375532cm) {$T_5$};
    \node[draw,circle,fill=green] at (1.58cm,3.0cm) {$T_6$};
    \node[draw,circle,fill=green] at (2.48214285714cm,4.82135155949cm) {$T_7$};
    \node[draw,circle,fill=green] at (2.47010519172cm,1.17638276213cm) {$T_8$};
    \node[draw,circle,fill=green] at (6.59261704682cm,4.49061910478cm) {$T_9$};
    \node[draw,circle,fill=green] at (6.59407518797cm,1.50546079484cm) {$T_{10}$};
    \node[draw,circle,fill=red] at (8.1cm,2.94693877551cm) {$V_1$};
    \node[draw,circle,fill=red] at (4.64285714286cm,0.4cm) {$V_2$};
    \node[draw,circle,fill=red] at (4.64285714286cm,5.6cm) {$V_3$};
    \node[draw,circle,fill=red] at (0.557142857143cm,4.22040816327cm) {$V_4$};
    \node[draw,circle,fill=red] at (0.557142857143cm,1.77959183673cm) {$V_5$};
    \node[draw,circle,fill=red] at (4.32857142857cm,4.32653061224cm) {$V_6$};
    \node[draw,circle,fill=red] at (4.32857142857cm,1.67346938776cm) {$V_7$};
    \node[draw,circle,fill=red] at (6.37142857143cm,3.26530612245cm) {$V_8$};
    \node[draw,circle,fill=red] at (2.91428571429cm,2.31020408163cm) {$V_9$};
    \node[draw,circle,fill=red] at (2.91428571429cm,3.68979591837cm) {$V_{10}$};
  \end{tikzpicture}
  \caption{
    Training and validation set pattern for the position of the coffee machine,
    designed to evaluate the generalization on a target area.
    The training pattern was selected with Lloyd's algorithm to cover the
    target area evenly, and is depicted with green circles. The evaluation
    pattern is depicted in red, and was selected to be far from the training
    points while covering evenly the target area. The numbers in the green
    and red circles represent the order used for training and validation
    positions for the coffee machine respectively.
  }
  \label{fig:coffee:pattern}
\end{figure}

A coffee preparation task was one of the tasks used to evaluate the proposed methods. We use
an inexpensive coffee grinder and an Aeropress as a brewing method. Figure~\ref{fig:coffee}
depicts the robot executing the steps required to prepare a cup of coffee. First, the
robot needs to move to the top of the grinder and pour fresh coffee beans. Subsequently,
the robot moves to place the spoon under the grinder funnel to pick the coffee grounds.
Finally, the robot pours the coffee grounds into the brewing chamber.

The coffee task requires sequencing movement primitives to pour coffee beans of grounds in 
different locations and picking the grounds from the grinder. At the same time, the robot should 
avoid hitting the grinder, coffee machine or the table to prevent damaging the robot,
the coffee machines or spilling the coffee. Therefore, this task allows us to test the
ability of the proposed framework to divide a complex task into multiple simpler
primitives as well as learning from the teacher the right set of movements that avoid 
hitting external objects.

Additionally, the movement primitives to pour or pick coffee should be adapted to
the position of the coffee machine or the grinder in order to succeed. The position
of these objects is obtained from the vision system in task space, providing an
opportunity to test the operator to adapt movement primitives in task space. The
operator to condition movement primitives in joint space is also used to start
the executed movement primitive at the robot current joint position.

For the coffee task we want to test how well the proposed approach adapts to
changes in the position of the grinder or the coffee machine, whereas for
the table tennis task the goal is to determine how well it adapts to changes
in the ball trajectory. Note that the position of the grinder and coffee machine
in different experiment trials is easy to control with relatively good precision,
while controlling the table tennis ball trajectory between different experiment
trials is virtually impossible. As a result, we decided to invest more effort
in the experiment design to test the generalization ability of a single movement
primitive in the coffee task.

\begin{algorithm}[t]
  \begin{algorithmic}[1]
    \Require Training set positions~$\{T_{1},\dots,T_{10}\}$ and validation set
      positions~$\{V_{1},\dots,V_{10}\}$ from Figure~\ref{fig:coffee:pattern}.
    \Ensure Training set performance~$P^t_n$ and evaluation set performance~$P^v_n$
     with~$n$ training samples for~$n \in \{1,\dots,10\}$.
    \For{ $n \in \{1,\dots,10\}$ }
      \State $\train(\{T_{1},\dots,T_{n}\})$
      \State $P^t_n \gets \evaluate(\{T_{1},\dots,T_{n}\})$
      \State $P^v_n \gets \evaluate(\{V_{1},\dots,V_{10}\})$
    \EndFor
  \end{algorithmic}
  \caption{Procedure to test the generalization performance of a single ProMP on a
  pouring coffee experiment}
  \label{alg:coffee:exp}
\end{algorithm}

To test the generalization ability of a single movement primitive we focused
on the movement that pours the coffee grounds in the coffee machine. We generated
a pattern with training and evaluation positions for the coffee machine with a
rectangular shape of 42cm x 59.4cm. This size corresponds exactly to an A2
format paper size that was printed for the experiments. 
To select a set of positions that covers evenly the training area we used Lloyd's
algorithm~\cite{lloyd1982least}. For the evaluation set we used an algorithm that selected
a set of points in the rectangle that maximized the distance to the training set. 
Figure~\ref{fig:coffee:pattern} shows a resized version of the resulting format 
for the training and the validation positions for the coffee machine as green and 
red circles respectively. The numbers on
the circles represent the order of the events that should be used in the 
experiment, and we used them to test the performance on the training and 
validation sets as a function of the training data. In this experiment we
evaluated the success rate of pouring coffee in the machine with a number of
training instances varying from 1 to 10 training samples. 

The procedure to train
and evaluate the performance is explained with detail as a pseudo-code
in Algorithm~\ref{alg:coffee:exp}. In this pseudo-code the~$\train(\cdot)$ function
consists on the human training the robot to pour coffee grounds on the coffee
machine on the positions passed as argument, and the~$\evaluate(\cdot)$ function
consists on the robot attempting to pour coffee on the specified positions
and evaluating the success rate. 
For example, to evaluate the validation set performance with two training examples, 
we would train the robot to pour coffee on positions~$\{T_1, T_2\}$ and subsequently 
evaluate the pouring performance in positions~$\{V_1,\dots,V_{10}\}$.

\rev{
  We used coffee beans instead of coffee grounds on the pouring experiments
  to simplify the definition of success in a pouring attempt. A trial is
  considered successful if and only if all the beans end up in the brewing 
  chamber after pouring. No spilling is allowed, a trial is considered failed if
  one or more beans fall out of the brewing chamber after the pouring movement
  is executed.
}

\begin{table}
  \centering
  \begin{tabular}{@{}ccc@{}}
    \toprule
    \bf{Training Samples} & \bf{Training} & \bf{Validation} \\
    \midrule
    1 & \rev{1/1} & \rev{1/10} \\
    \{2,\dots,10\} & \rev{10/10} & \rev{10/10} \\
    \bottomrule
  \end{tabular}
  \caption{
    Summary of the results of the generalization performance experiment
    pouring coffee with a single ProMP. Using only two training samples
    was enough to generalize to all the target area. With one training 
    sample (T1), the robot succeeded only for the provided training
    point (T1) and the closest of the validation points (V5), spilling
    coffee in the rest of the evaluated positions. The obtained results
    suggest that at least for this task the selected prior is a sensible
    choice.
  }
  \label{tab:coffee:results}
\end{table}

Table~\ref{tab:coffee:results} summarize the results of the pouring performance
measured on this experiment. We expected a curve of generalization performance
increasing slowly as a function of the number of training data, but the results
obtained showed that after demonstrating the pouring movement only in~$T_1$ and~$T_2$
the robot could successfully generalize to all the \rev{validation points}. 
\rev{
With the same two training instances we tried to validate generalization in the 
points~$\{T_3,\dots,T_{10}\}$ and the robot successfully poured coffee in those 
positions as well. Note that the results presented in Table~\ref{tab:coffee:results}
do not mean that the presented approach can generalize to any pouring point given
only two demonstrations. If for example, we provide~$T_1$ and~$T_8$ as training
examples and attempted to validate in the rest of the pouring area, not only the
pouring is likely to fail but the resulting planned movement might be dangerous
to execute. The ProMPs, as most machine learning methods that assume independently
identically distributed data (IID), does not handle extrapolation well. The ProMP
framework could be extended using transfer learning techniques~\cite{pan2010survey}
to deal better with a non IID scenario, but such an extension is outside the scope
of this paper.
}

With only one training instance of pouring the robot could not generalize well.
However, note that the robot managed to pour successfully at the given training
position and one of the validation positions. The validation position where the
robot poured successfully was~$V_5$, that is the closest validation point to the
given training point~$T_1$, as can be seen in Figure~\ref{fig:coffee:pattern}.
The distance between~$T_1$ and~$V_5$ in the printed pattern is 10.4 cm. 
We also tried to validate the single training instance example on the 
points~$\{T_2,\dots,T_{10}\}$, but it failed spilling the coffee every time.


An alternative method to solve the coffee task without learning from human 
demonstrations would require trajectory planning with collision avoidance 
in order to succeed. 
Additionally, common sense
knowledge like keeping the spoon pointing up all the time except when the
robot is pouring would have to be explicitly programmed.
Instead, our approach learns these common sense knowledge and strategies
to succeed avoiding collisions with the grinder and brewing chamber from 
the human demonstrations.
In the next section
we evaluate our method in a table tennis task. We believe that robot table 
tennis is significantly harder than the coffee task presented in this section
for a number of reasons that we discuss with more detail in the following section.
Unfortunately, it is very hard to control precisely the ball trajectory and
as a result, we cannot provide detailed generalization performance as with
the coffee task. Instead, we will focus on evaluating the hit and return rate
performance compared to previous work.

\subsection{Robot Table Tennis}




Robot table tennis is a highly dynamic task difficult to play for robots
and humans. Unlike the coffee task it has strong real time requirements.
The timing of the movement is as important as the movement itself to
succeed hitting and returning the ball. Furthermore, it is not trivial 
or obvious which kind on movements
would result in success for a given ball trajectory, making this problem
especially interesting for learning approaches that can uncover these 
patterns given a set of successful trial examples.

In this section we evaluate the proposed approach in a robot table tennis
setup. In this task we use a table tennis ball gun to throw balls to the 
robot. Subsequently, we measure weather or not the robot hits
the ball and if the ball landed successfully in the opponent's court 
according to the table tennis rules. 

For all the experiments presented
in this section, we collected eight human demonstrations of a particular
striking movement to train a ProMP. Unlike the coffee task, the high
variability in the results makes it hard to determine the optimal number
of training samples to increase the success rate. Informally, we
did not notice any significant performance improvements using more
than eight demonstrations.

To segment the striking movement from
the rest of the demonstrated behavior we used the zero crossing velocity
heuristic method. First, we found the point where the racket hit the ball~$t_h$
by detecting the change in direction of the ball. Subsequently, we found
a time interval~$(t_a, t_b)$ such that~$t_h \in (t_a,t_b)$ and both~$t_a$
and~$t_b$ were zero crossing velocity points. We found that this heuristic
reliably segments table tennis striking movements if the hitting 
time~$t_h$ can be detected accurately. Some times we could not detect
the hitting time~$t_h$ accurately because of vision problems. In such case
we simply discarded that trajectory from the training set. We decided to use
six as the minimum number of segmented demonstrations in the training set to 
proceed with the experiments. That is, if more than two demonstrations were
discarded by the segmentation heuristic we collected the training data again.


Let us explain in detail how we apply the proposed method to table 
tennis as well the similarities and differences to previous work presented
in~\cite{gomez2016using}. A high level pseudo-code of the table tennis
strategy is presented in Algorithm~\ref{alg:tt:proc}. This algorithm receives as input
a ProMP already trained to play table tennis using human demonstrations, moves
the robot to an initial position and blocks its execution until the vision
system produces new ball observations. Subsequently, the obtained ball
observations are used to predict the rest of the ball trajectory using
a Kalman Filter, the optimal initial time is computed from the ball
trajectory using a maximum likelihood approach introduced in previous
work~\cite{gomez2016using}, and the trained ProMP is conditioned in
task space using the operator presented in Section~\ref{sec:promp:kinematics}.
Before executing the ProMP conditioned to hit the ball, it is conditioned
in joint space to start in the current robot joint state.

\begin{algorithm}[t]
  \begin{algorithmic}[1]
    \Require A ProMP~$\text{promp0}$ trained for table tennis using
      human demonstrations.
    \While{ running }
      \State $\text{move\_to\_init\_state}(\text{promp0})$
        \label{alg:tt:init_state}
      \State $\text{wait\_ball\_obs}()$
        \label{alg:tt:proc:wait_ball_obs}
      \Repeat
        \State $\text{ball\_obs} \gets \text{get\_ball\_obs}()$
          \label{alg:tt:proc:get_ball_obs}
        \State $\text{ball\_traj} \gets \text{predict\_ball\_traj}(\text{ball\_obs})$
          \label{alg:tt:proc:pred_ball_traj}
        \State $t_0 \gets \text{comp\_optimal\_t0}(\text{ball\_traj})$
          \label{alg:tt:proc:comp_opt_t0}
        \State $\text{new\_promp} \gets \text{cond\_hit}(\text{promp0},\text{ball\_traj})$
          \label{alg:tt:proc:cond_promp_hit}
      \Until{ $t_0 \ge \text{current\_time}()$ }
      \State $\text{new\_promp.cond\_joint\_space}(\text{get\_joint\_state}())$
        \label{alg:tt:proc:cond_joint_space}
      \State $\text{execute}(\text{new\_promp})$
        \label{alg:tt:proc:execute}
    \EndWhile
  \end{algorithmic}
  \caption{Procedure used on the table tennis experiments}
  \label{alg:tt:proc}
\end{algorithm}

Note that the lines~\ref{alg:tt:proc:get_ball_obs} and~\ref{alg:tt:proc:cond_promp_hit}
in Algorithm~\ref{alg:tt:proc} are in a loop to allow for re-planning. This feature
is an important improvement over the previous work presented in~\cite{gomez2016using},
because it allows for corrections over the predictions of the ball trajectory produced
by the Kalman Filter in line~\ref{alg:tt:proc:pred_ball_traj}.
In~\cite{gomez2016using}, a set of ball observations of a certain size was obtained
and the Kalman Filter was used only once to predict the rest of the ball trajectory.
Subsequently, the robot would ``close its eyes'' and attempt to hit the predicted
ball trajectory. In consequence, it was hard to fix a sensible size for the initial
set of observations. A small set would not provide enough information to predict
accurately the ball trajectory, and a large set could potentially leave a small
reaction time to the robot effectively loosing the opportunity to hit the ball.
In this paper, we took advantage of the short execution time of the presented
operators using re-planning. We simply take any amount of available ball
observations to predict the ball trajectory and adapt the ProMP, but we keep
doing so while there is still time for corrections.

The starting time of the movement primitive is computed in Line~\ref{alg:tt:proc:comp_opt_t0}
using the operator presented in~\cite{gomez2016using} that maximizes the likelihood of
hitting the ball under some assumptions. To compute this likelihood without specifying a
hitting time or point, the hitting time was marginalized using some prior distribution.
In~\cite{gomez2016using}, a uniform distribution was used as prior over the hitting time.
We observed that the human teachers usually hit the ball close to the middle of the movement.
In consequence, we changed the prior distribution over the hitting time to match
the observed teacher behavior. We used a Gaussian distribution given by
\[p_h(z) = \normal{z}{\mu_z=0.5}{\sigma_z=0.1}\]
where~$z=(t-t_0)/T$ is the time variable normalized to be between zero and one.
As a result, we obtained a substantial improvement on the number of times that
the robot manages to successfully return the ball to the opponent's court, that
we will call in the rest of this paper the \textit{success rate}.

\begin{table}
  \centering
  \begin{tabular}{@{}cccc@{}}
    \toprule
    \bf{Re-planning} & \bf{Hit time prior} & \bf{Hit rate} & \bf{Success rate} \\
    \midrule
    No & Uniform & 73.7\% & 5.2\% \\
    No & Gaussian & 79.5\% & 40.9\% \\
    Yes & Uniform & 93.2\% & 9.1\% \\
    Yes & Gaussian & 96.7\% & 67.7\% \\
    \bottomrule
  \end{tabular}
  \caption{
    Performance improvement for hit and success rate due to re-planning and the 
    prior over the hitting time. The ball gun was fixed to the same settings on
    these four experiments, and the same ProMP was used in every case trained with
    Algorithm~\ref{alg:em_promp}. The goal of these experiments was to test the
    effect of re-planning and the hitting time prior both independently and combined.
    Note that re-planning has a significant positive impact mostly over the hitting 
    rate, whereas the prior over the hitting time affects mostly the success rate.
    The best performance is obtained as expected with a combination of both.
  }
  \label{tab:tt:tech}
\end{table}

Replanning and the prior over the hitting time are features added to the table
tennis strategy on this paper that were not present in~\cite{gomez2016using}.
Although these features are unrelated to the main contributions of this paper,
we consider important to evaluate the performance improvement due to these
features to explain the huge performance gap in comparison to the performance
reported in~\cite{gomez2016using}. In addition, the replanning feature is
possible only because of the fast execution time of the proposed adaptation 
operators. Therefore, replanning is an example of how the computational
efficiency of the proposed methods can have an impact on the success of
a task where accurate prediction models are not available.

Table~\ref{tab:tt:tech} presents the results of an experiment to measure the
improvement of performance due to re-planning and the hitting time prior both
independently and combined. We placed the ball gun in a position that the human
teacher found comfortable and collected a set of demonstrations, the ball gun
parameters were kept fixed during the rest of the experiment. We trained a
ProMP with Algorithm~\ref{alg:em_promp} using the collected demonstrations.
We use the exact same trained ProMP during this experiment to make sure
that the measured improvements are only due to the re-planning and
hitting prior features. Note that the change in the prior over the hitting 
time had a very significant impact on the success rate, increasing it from
5.2\% to 40.9\% without re-planning and from 9.1\% to 67.7\% with
re-planning. On the other hand, the re-planning feature improved in general
about 20\% on the hit rate, and the success rate improvement was only substantial
in combination with the hitting time prior, improving from 40.9\% to 67.7\%.


A major difference between this work and~\cite{gomez2016using}, is the
training algorithm for the movement primitives. In~\cite{gomez2016using},
the movement primitives were trained with a maximum likelihood algorithm.
In Section~\ref{sec:methods}, we discussed how maximum likelihood
estimation (MLE) produced unstable estimates of the ProMP parameters
opposed to the Maximum A-Posteriori estimates (MAP). To prevent
stability problems, the MLE estimates computed in~\cite{gomez2016using}
force the matrix~$\matr{\Sigma}_\omega$ to be block diagonal. As
a result, the computed ProMP considers all the joints independent.

To measure the effect of using the proposed training method opposed to 
considering all the joints independent with MLE, we tested ProMPs
trained with both methods with several ball gun configurations using
always the procedure for execution on Algorithm~\ref{alg:tt:proc} with both
re-planning and the prior over the hitting time.
We obtained
an average success rate of 66.3\% and a hit rate of 95.4\% for 
the MAP trained ProMP. For MLE we obtained an average of 47.7\% 
and 79.8\% for the hit and success rates respectively.

\begin{figure}
  \centering
  \includegraphics[width=6cm]{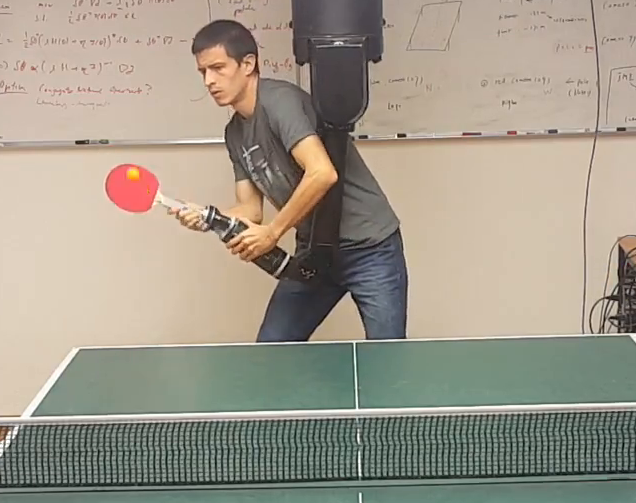}
  \caption{
    A human subject moving the robot in gravity compensation mode.
    Gravity compensation mode was used to obtain the human
    demonstrations necessary to train the robot. 
  }
  \label{fig:teach:fh}
\end{figure}

We also compare the performance of our method with a different
robot table tennis method based on heuristics~\cite{mulling2011biomimetic}
called the MoMP method~\cite{mulling2013learning}. 
Figure~\ref{fig:teach:fh} shows
a human subject moving the robot in gravity compensation mode. 

\begin{figure*}
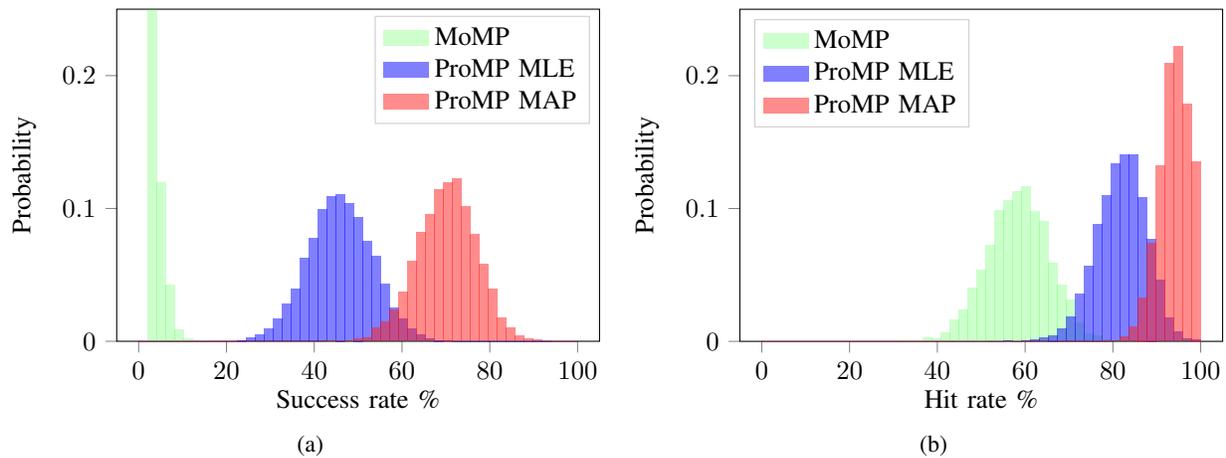

  \centering
  \setlength\figurewidth{8cm}
  \setlength\figureheight{6cm}
  \setlength\axislabelsep{0mm}
  \begin{subfigure}[b]{0.45\textwidth}
    \tikzsetnextfilename{success}
    \input{fig/success.tex}
    \caption{}
    \label{fig:tt:map_mle:success}
  \end{subfigure}
  \begin{subfigure}[b]{0.45\textwidth}
    \tikzsetnextfilename{hit}
    \input{fig/hit.tex}
    \caption{}
    \label{fig:tt:map_mle:hit}
  \end{subfigure}
  \caption{
    Histogram of the success and hit rates on table tennis for ProMPs trained with MAP 
    and MLE. We also compare against humans moving the robot in gravity compensation mode
    and the MoMP method. For the table tennis experiments, repeating the same experiment
    multiple times will likely produce different hit and success rate performance.
    In consequence, we decided to present a histogram of the results computed with the
    bootstrap method representing how likely is it to obtain a particular hit or success 
    rate for different methods.
  }
  \label{fig:tt:map_mle}
\end{figure*}

Figure~\ref{fig:tt:map_mle} shows an histogram of the success and hit rates
obtained in this experiment for both MAP and MLE training, the MoMP method
and the human subjects. The histogram
was generated with the bootstrap method, generating 5000 random samples
of 50 trials from the collected data. The success and hit rates were computed
for each of the 5000 samples and recorded in the histogram. 
We decided to present an histogram of these results instead of just a number
to account for the variability of the results natural to the table tennis experiments. 

An interval containing 90\% of the probability mass of the success rate histogram
for the MLE and MAP trained ProMPs would locate the success rate between 34.0\% and 58.0\%
for MLE and between 60.0\% and 80.0\% for MAP. From these confidence intervals we
can conclude that the difference in success rate of learning the joint correlations 
with the MAP algorithm presented on this paper compared to the MLE algorithm
presented in~\cite{gomez2016using} that assumes the joints as independent is
significant. 

Furthermore, the table tennis procedure presented in 
Algorithm~\ref{alg:tt:proc} used for the MAP and MLE trained ProMPs does
not include any heuristic or method to successfully return the ball to the
opponent's court. In both cases this behavior has to be learned from the
demonstrated data. The fact that the success rate of the MAP trained ProMP
is significantly better than the success rate of the MLE trained ProMP
that forces~$\matr{\Sigma}_\omega$ to be block diagonal, suggests
that the joint correlations encode information important to successfully
return table tennis balls.


The performance of the presented approach was significantly better than the
MoMP method for both hit and success rates in our experiments. The MoMP
method is based on several heuristics that would require a great
amount of hand tuning to achieve a good success rate for a particular ball
gun configuration. As a result, it is very hard to tune this method to
generalize well to different ball gun locations and orientations.
On the other hand, our method generalizes well to changes on the ball
trajectory and can be easily retrained if the ball gun configuration
is significantly changed.

\section{Conclusions and Discussion}

This paper introduces new operators to learn and adapt probabilistic movement primitives
in joint and in task space. The presented learning algorithm uses a prior distribution 
to increase the robustness of the estimated parameters. Using the proposed prior distribution 
over the ProMP parameters is an effective way to improve robustness and learn with few training
instances while conserving enough flexibility in the model to learn the dependencies
between the joints as more data becomes available. 

This paper also presents simple and fast operators to adapt movement primitives in joint and 
task space, making use of standard methods of probability theory. These operators were
evaluated in the coffee task and table tennis task to adapt the learned movement primitive
to the coffee machine position and the ball trajectory respectively. The presented operators 
to adapt movement primitives can be applied to any other robotic applications.


We have compared the table tennis performance of the presented approach with previous work
presented in~\cite{gomez2016using}. We tested the performance improvements due to 
table tennis specific advantages like re-planning and the prior over the 
hitting time. More importantly, we tested the improvements due to the
presented learning algorithm and its ability to learn the joint correlations 
independently of the table tennis specific improvements. We show that the difference
on the learning algorithm alone is enough to obtain a statistically significant
improvement.

Unlike previous approaches to robot table tennis, our approach does not model the interaction 
between the racket and the ball. The reason why the presented method can successfully 
send balls to the opponent's side of the table is because the training data used to learn
the movement primitive contains mostly successful examples. Thus, the behavior of 
successfully returning balls is completely learned from data.

A limitation of the presented training method is that it requires manual segmentation
of the robot trajectories. Someone needs to specify where every movement primitive
starts and ends in the demonstrated behavior. A better approach would be to consider
the segmentation as another hidden variable and add it to the proposed EM inference
algorithm. The problems of automatic segmentation and clustering should be considered 
in future work.



%

\appendices

\ifCLASSOPTIONcaptionsoff
  \newpage
\fi



\bibliographystyle{IEEEtran}
\bibliography{IEEEabrv,refs}
%



%

\begin{IEEEbiography}[{\includegraphics[width=1in,height=1.25in,clip,keepaspectratio]{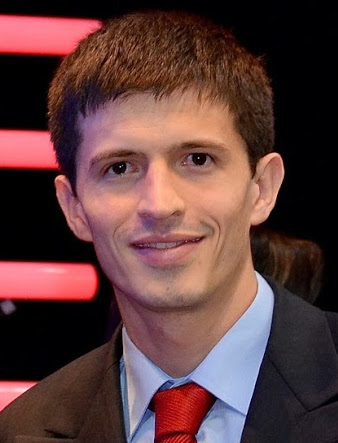}}]{Sebastian Gomez-Gonzalez}
  joined the MPI for Intelligent Systems in 2015. He is 
  also affiliated with Technische Universitaet Darmstadt as an external member. His
  research interests include machine learning, generative models for motion and
  reinforcement learning.
  Sebastian received his MSc and BSc degree in computer science from Universidad 
  Tecnologica de Pereira. He obtained the ``Best in Education'' award for obtaining
  the best score in Colombia in the standardized test for computer science in 2011.
\end{IEEEbiography}

\begin{IEEEbiography}[{\includegraphics[width=1in,height=1.25in,clip,keepaspectratio]{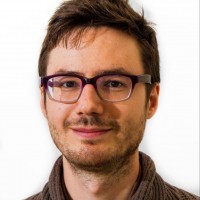}}]{Gerhard Neumann}
  is a Professor of Robotics \& Autonomous Systems in College of Science of Lincoln University. 
  Before coming to Lincoln, he has been an Assistant Professor at the TU Darmstadt. 
  Before that, he was Post-Doc and Group Leader at the 
  Intelligent Autonomous Systems Group (IAS) also in Darmstadt. 
  Gerhard obtained his Ph.D. under the supervision of Prof. Wolfgang Mass at the 
  Graz University of Technology.
  Gerhard has a strong publication record both in machine learning venues 
  (e.g., NIPS, ICML) and in the robotics community (e.g., ICRA, IROS). He 
  has been active at bringing researchers from both fields together by 
  organizing multiple workshops at the frontier between these two fields, 
  e.g., on reinforcement learning and motor skill acquisition. He served 
  in the senior program committee of some of the most prestigious 
  conferences in artificial intelligence including NIPS and AAAI.
\end{IEEEbiography}

\begin{IEEEbiography}[{\includegraphics[width=1in,height=1.25in,clip,keepaspectratio]{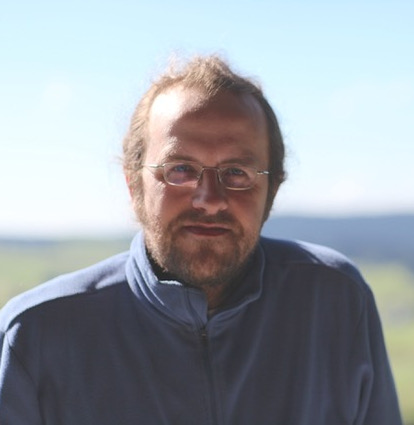}}]{Bernhard Schölkopf}
  's scientific interests are in machine learning and causal inference. 
  He has applied his methods to a number of different application areas, ranging from 
  biomedical problems to computational photography and astronomy. Bernhard has researched 
  at AT\&T Bell Labs, at GMD FIRST, Berlin, and at Microsoft Research Cambridge, UK, before 
  becoming a Max Planck director in 2001. He is a member of the German Academy of 
  Sciences (Leopoldina), and has received the J.K. Aggarwal Prize of the International 
  Association for Pattern Recognition, the Max Planck Research Award (shared with S. 
  Thrun), the Academy Prize of the Berlin-Brandenburg Academy of Sciences and 
  Humanities, and the Royal Society Milner Award.
\end{IEEEbiography}

\begin{IEEEbiography}[{\includegraphics[width=1in,height=1.25in,clip,keepaspectratio]{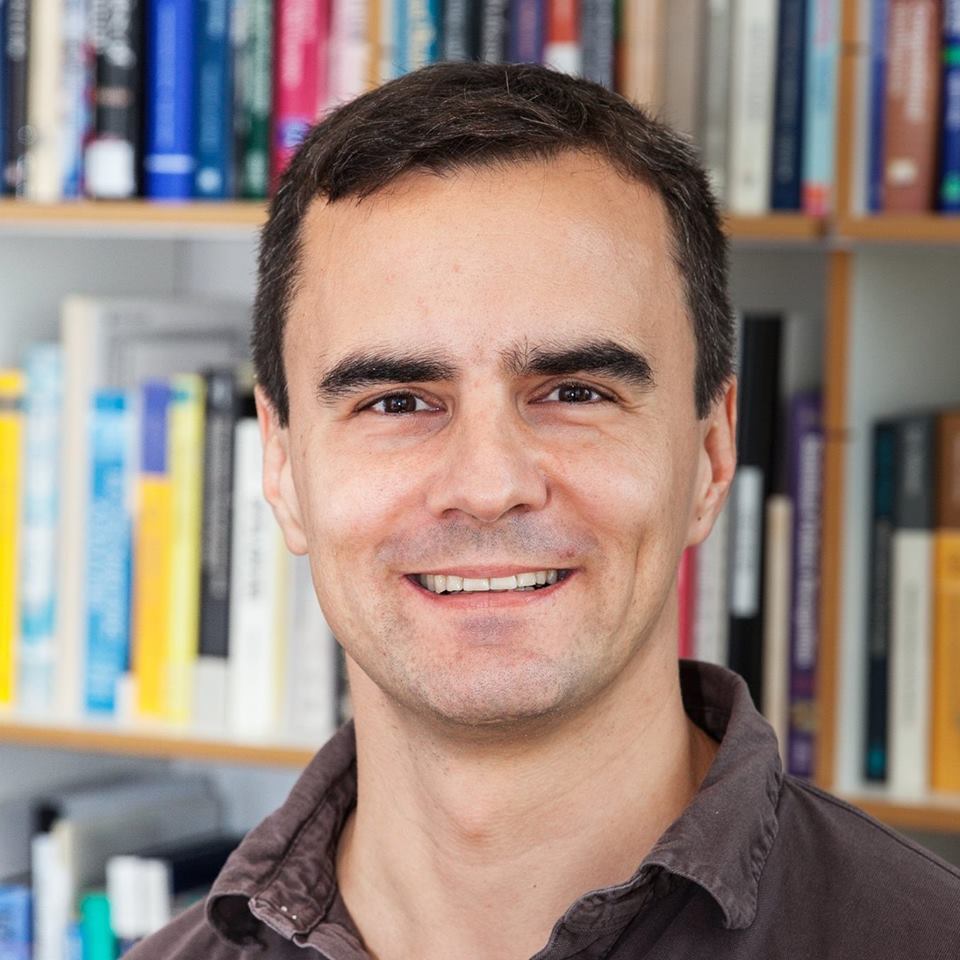}}]{Jan Peters}
  is a full professor (W3) for Intelligent Autonomous Systems at the
  Computer Science Department of the Technische Universitaet Darmstadt and at 
  the same time a senior research scientist and group leader at the Max-Planck 
  Institute for Intelligent Systems, where he heads the interdepartmental Robot 
  Learning Group. Jan Peters has received the Dick Volz Best 2007 US PhD Thesis 
  Runner-Up Award, the Robotics: Science \& Systems - Early Career Spotlight, the 
  INNS Young Investigator Award, and the IEEE Robotics \& Automation Society's 
  Early Career Award. Recently, he received an ERC Starting Grant.
  In 2019, Jan Peters was appointed IEEE Fellow.
  Jan Peters has studied Computer Science, Electrical, Mechanical and Control 
  Engineering at TU Munich and FernUni Hagen in Germany, at the National University 
  of Singapore (NUS) and the University of Southern California (USC). He has received 
  four Master's degrees in these disciplines as well as a Computer Science PhD from USC. 
  Jan Peters has performed research in Germany at DLR, TU Munich and the Max Planck 
  Institute for Biological Cybernetics (in addition to the institutions above), in 
  Japan at the Advanced Telecommunication Research Center (ATR), at USC and at both 
  NUS and Siemens Advanced Engineering in Singapore.
\end{IEEEbiography}





\end{document}